\documentclass[journal,doublecolumn]{IEEEtran}
\usepackage{amsmath,graphicx}
\usepackage[english]{babel}
\usepackage[T1]{fontenc}
\usepackage{url}
\usepackage{graphicx}
\usepackage{subfigure}
\usepackage{subfig}  
\usepackage{alphalph} 
\usepackage{epstopdf}
\usepackage{epsfig}

\usepackage{color}
\usepackage{amsmath,bm}
\usepackage{amsthm,amsfonts,amssymb}

\usepackage{booktabs}
\usepackage{colortbl}
\usepackage{pstricks}
\usepackage{pst-node}
\usepackage{pst-grad}
\usepackage{pst-plot}
\usepackage{pstricks-add}
\usepackage{ifthen}
\usepackage{algorithm}
\usepackage{algorithmic}	
\usepackage{float}
\usepackage{fp}
\usepackage{cite}
\usepackage{enumerate}
\usepackage{multicol} 

\usepackage{array}
\newcolumntype{L}[1]{>{\raggedright\arraybackslash}m{#1}}
\newcolumntype{C}[1]{>{\centering\arraybackslash}m{#1}}
\newcolumntype{R}[1]{>{\raggedleft\arraybackslash}m{#1}}

\newfloat{algorithm}{t}{lop}

\newcommand{\mypar}[1]{{\bf #1.}}

\theoremstyle{plain}

\definecolor{red}{RGB}{153,0,0}
\definecolor{green}{RGB}{0,153,0}			
\definecolor{blue}{RGB}{0,0,153}
\definecolor{darkred}{RGB}{90,0,0}
\definecolor{darkgreen}{RGB}{0,90,0}
\definecolor{darkblue}{RGB}{0,0,90}

\title{Multimodal Image Super-resolution via Joint Sparse Representations induced by Coupled Dictionaries}
\author{Pingfan~Song, \textit{Student Member, IEEE},
	Xin Deng, \textit{Student Member, IEEE}, \\
	Jo\~ao F.\ C.\ Mota, \textit{Member, IEEE},
	Nikos Deligiannis, \textit{Member, IEEE}, \\
	Pier Luigi Dragotti, \textit{Fellow, IEEE}, and
	Miguel R.\ D.\ Rodrigues, \textit{Senior Member, IEEE}
	\thanks{This work is supported by China Scholarship Council (CSC), UCL Overseas Research Scholarship (UCL-ORS), the VUB-UGent-UCL-Duke International Joint Research Group grant, and by EPSRC grant EP/K033166/1.}
	\thanks{Pingfan Song and Miguel R.\ D.\ Rodrigues are with the Department of Electronic \& Electrical Engineering, University College London, London WC1E 6BT, UK. (e-mail: pingfan.song.14@ucl.ac.uk, m.rodrigues@ucl.ac.uk)}
	\thanks{Xin Deng and Pier Luigi Dragotti are with the Department of Electronic \& Electrical Engineering, Imperial College London, London SW7-2AZ, UK. (e-mail: x.deng16@imperial.ac.uk, p.dragotti@imperial.ac.uk)}
	\thanks{Jo\~ao F.\ C.\ Mota is with the Department of School of Engineering \& Physical Sciences, Heriot-Watt University, Edinburgh EH14 4AS, UK. (email: j.mota@hw.ac.uk)}
	\thanks{N. Deligiannis is with the Department of Electronics and Informatics, Vrije Universiteit Brussel, B-1050 Brussels, Belgium, and with imec, Kapeldreef 75, B-3001 Leuven, Belgium. (e-mail: ndeligia@etrovub.be)}%
	}

\begin{document}
%
\maketitle

\begin{abstract}
	Real-world data processing problems often involve various image modalities associated with a certain scene, including RGB images, infrared images or multi-spectral images. The fact that different image modalities often share certain attributes, such as certain edges, textures and other structure primitives, represents an opportunity to enhance various image processing tasks. This paper proposes a new approach to construct a high-resolution (HR) version of a low-resolution (LR) image given another HR image modality as reference, based on joint sparse representations induced by coupled dictionaries. Our approach, which captures the similarities and disparities between different image modalities in a learned sparse feature domain in \emph{lieu} of the original image domain, consists of two phases. The coupled dictionary learning phase is used to learn a set of dictionaries that couple different image modalities in the sparse feature domain given a set of training data. In turn, the coupled super-resolution phase leverages such coupled dictionaries to construct a HR version of the LR target image given another related image modality. One of the merits of our sparsity-driven approach relates to the fact that it overcomes drawbacks such as the texture copying artifacts commonly resulting from inconsistency between the guidance and target images. Experiments on real multimodal images demonstrate that incorporating appropriate guidance information via joint sparse representation induced by coupled dictionary learning brings notable benefits in the super-resolution task with respect to the state-of-the-art. Of particular relevance, the proposed approach also demonstrates better robustness than competing deep-learning-based methods in the presence of noise.
\end{abstract}




\section{Introduction}


\label{sec:intro}
Image super-resolution (SR) is an operation that involves the enhancement of pixel-based image resolution, while minimizing visual artifacts. 
However, the construction of a high-resolution (HR) version of a low-resolution (LR) image requires inferring the values of missing pixels, making image SR a severely ill-posed problem. Various image models and approaches have been proposed to regularize this ill-posed problem via employing some prior knowledge, including natural priors~\cite{li2001new,dai2007soft,sun2008image,zhang2008image}, local and non-local similarity~\cite{yang2013fast,dong2013nonlocally}, sparse representation over fixed or learned dictionaries~\cite{mallat2010super, yang2010image,yang2012coupled,zeyde2010single,timofte2013anchored,wei2016fresh,timofte2014a+}, and sophisticated features from deep learning~\cite{dong2016image,dong2016accelerating,kim2016accurate,kim2016deeply}. These typical super-resolution approaches focus only on single modality images without exploiting the availability of other modalities as guidance.

However, in many practical application scenarios, a certain scene is often imaged using different sensors yielding different image modalities. For example, in remote sensing it is typical to have various image modalities of earth observations, such as a panchromatic band version, a multi-spectral bands version, and an infrared (IR) band version~\cite{gomez2015multimodal,loncan2015hyperspectral}. In order to balance cost, bandwidth and complexity, these multimodal images are usually acquired with different resolutions~\cite{gomez2015multimodal}. 
%
These scenarios call for approaches that can capitalize on the availability of multiple image modalities of the same scene -- which typically share textures, edges, corners, boundaries, or other salient features -- in order to super-resolve the LR images with the aid of the HR images of a different modality. 

Therefore, a variety of joint super-resolution/upsampling approaches have been proposed to leverage the availability of additional \emph{guidance images}, also referred to as \emph{side information}\cite{renna2016classification,mota2017compressed}, to aid the super-resolution of target LR modalities\cite{kopf2007joint,he2010guided,ham2017robust,li2016deep,shen2015multispectral,zhang2014rolling}. The basic idea behind these methods is that the structural details of the guidance image can be transferred to the target image. However, these methods tend to introduce notable texture-copying artifacts, i.e. erroneous structure details that are not originally present in the target image because such methods typically fail to distinguish similarities and disparities between the different image modalities. 

The motivation of this work is to introduce a new image SR approach, based on joint sparse representations induced by coupled dictionaries, that has the ability to take into account both similarities and disparities between target and guidance images in order to deliver superior SR performance.

{\bf Proposed Scheme}.
The proposed scheme is based on three elements: (1) a data model; (2) a coupled dictionary learning algorithm; and (3) a coupled image super-resolution algorithm.

\begin{itemize}
	\item 
	\emph{Data Model}: 
	This is a patch-based model that relies on the use of coupled dictionaries to jointly sparsely represent a pair of patches from the different image modalities. Of particular relevance is the ability to represent the similarities and disparities between the different image modalities in this sparse feature domain in \emph{lieu} of the original image domain, which leads to a higher super-resolution accuracy.
	
	\item 
	\emph{Coupled Dictionary Learning}: This algorithm learns the data model -- including a set of coupled dictionaries along with the joint sparse representations of the different image modalities -- from a set of training images.
	
	\item 
	\emph{Coupled Image Super-Resolution}: This algorithm uses the learned coupled dictionaries to perform joint sparse coding for the target/guidance image pair. The resulting joint sparse representations are then used to estimate a HR version of the target image from its LR version.
	
\end{itemize}

\noindent
In comparison with state-of-the-art approaches~\cite{kopf2007joint,he2010guided,ham2017robust,li2016deep,shen2015multispectral}, our approach can better model the common and distinct features of the different data modalities. This capability makes our approach more robust to inconsistencies between the guidance and the target images, as both the target LR image and the guidance image are taken into account during the estimation of the target HR image, instead of unilaterally transferring the structure details from the HR guidance image. 
In addition, our approach is also more robust to mismatches between training data and testing data (e.g. due to the presence of noise) in comparison to deep-learning-based approaches~\cite{li2016deep}.
%
%


\mypar{Contributions}
Our contributions are as follows:
\begin{itemize}
	\item
	We devise a data model for multimodal signals that captures the similarities and disparities between different modalities using joint sparse representations induced by coupled dictionaries. Compared with our previous work~\cite{song2016coupled}, the present model is more general, because it does not require the matrix that models the conversion of a HR version of the image to the LR counterpart to be known.
	
	\item
	We also propose a learning algorithm to learn the coupled dictionaries from different data modalities. Again, compared with~\cite{song2016coupled}, in the learning stage, the proposed algorithm does not require the knowledge of the matrix that converts a HR image to a LR version.
	
	\item
	We also propose a multimodal image super-resolution algorithm that enhances the resolution of the target LR image with the aid of another guidance HR image modality.
	
	\item
	Finally, extensive experiments are conducted both on a variety of multimodal images. 
	The results demonstrate that our proposed approach leads to better super-resolution performance than state-of-the-art approaches in a range of scenarios.

\end{itemize}

\mypar{Organization}
The remainder of this paper is organized as follows. We review related work in Section \ref{sec:RelatedWork}, including single and joint image SR, as well as other multimodal image processing works. We then propose our multimodal image super-resolution framework, including the data model, the coupled dictionary learning algorithm, and the multimodal image super-resolution algorithm in Section \ref{sec:SIMIS}.
Section \ref{sec:Experiments} is devoted to various simulation and practical experiments which demonstrate that our approach can lead to significant gains over the state of the art. We summarize the main contributions of the paper in Section V.

\section{Related Work}
\label{sec:RelatedWork}

There are various image super-resolution approaches in the literature. 
Single image super-resolution approaches do not leverage other guidance images, whereas joint image super-resolution approaches explicitly leverage the availability of other image modalities.


\vspace{-0.2cm}

\subsection{Single image SR}
In general, conventional single image SR approaches can be categorized into three classes: (1) interpolation-based, (2) reconstruction-based and (3) learning-based SR approaches.

\mypar{Interpolation-based SR approaches}
Advanced interpolation approaches exploit natural image priors, such as edges~\cite{li2001new}, image smoothness~\cite{dai2007soft}, gradient profile~\cite{sun2008image} and other geometric regularity of image structures~\cite{zhang2008image}. These methods are simple and fast, but tend to overly smooth image edges and generate ringing and jagged artifacts. 



\mypar{Reconstruction-based SR approaches}
Reconstruction-based SR approaches, also referred to as model-based SR methods, attempt to regularize the highly under-determined image SR inverse problem by exploiting various image priors, including self-similarity of images patches\cite{yang2013fast}, sparsity in the wavelet domain~\cite{mallat2010super}, analysis operator~\cite{hawe2013analysis}, and other fused versions\cite{dong2013nonlocally}. Recent work\cite{wei2016fresh} proposes a piecewise smooth image model and makes use of the finite rate of innovation (FRI) theory to reconstruct HR target images. These reconstruction-based methods usually offer better performance than interpolation-based methods.


\mypar{Learning-based SR approaches}
These SR approaches typically consist of two phases: (1) a learning phase where one learns certain image priors from training images and (2) a testing phase where one obtains the HR image from the LR version with the aid of the prior knowledge. 

In particular, patch-wise learning-based approaches leverage learned mappings or co-occurrence priors between LR and HR training image patches to predict the fine details in the testing target HR images according to their corresponding LR versions~\cite{bevilacqua2012low, yang2008image, yang2010image, yang2012coupled, zeyde2010single, timofte2013anchored, timofte2014a+, wang2012semi, jia2013image}.
%
%
%
For example, motivated by Compressive Sensing~\cite{donoho2006compressed,candes2006robust}, Yang \textit{et al.}~\cite{yang2008image, yang2010image, yang2012coupled} propose a sparse-coding based image SR strategy, which is improved further by Zeyde, \textit{et al.}\cite{zeyde2010single}. The key idea is a sparse representation invariance assumption which states that HR/LR image pairs share the same sparse coefficients with respect to a pair of HR and LR dictionaries. 
Along similar lines, Timofte \textit{et al.}~\cite{timofte2013anchored,timofte2014a+} propose a strategy, referred to as anchored neighbourhood regression, that combines the advantage of neighbor embedding and dictionary learning.
%
In order to achieve better flexibility and stability of signal recovery, semi-coupled dictionary learning\cite{wang2012semi} and coupled dictionary learning \cite{jia2013image} are proposed to relax the sparse representation invariance assumption to the same support assumption, allowing more flexible mappings.
Note that, even though the terminology related to "coupled dictionary learning" also appears in these works\cite{yang2012coupled,wang2012semi,jia2013image}, their approaches focus only on coupling LR and HR images of the same modality, and do not take advantage of other image modalities. In addition, their assumptions, models and algorithms are also different from ours.

Inspired by sparse-coding-based SR methods, Dong \textit{et al.}\cite{dong2016image} propose a single image super-resolution convolutional neural network (SRCNN) consisting of a patch extraction and representation layer, a non-linear mapping layer and a reconstruction layer. A faster and deeper version FSRCNN was proposed in~\cite{dong2016accelerating}, where the previous interpolation operation is removed and a deconvolution layer is introduced at the end of the network to perform upsampling. Kim \textit{et al.}\cite{kim2016accurate} propose a very deep SR network (VDSR) which exploits residual-learning for fast converging and multi-scale training datasets for handling multiple scale factors. Different from the above CNN-based SR approaches, \cite{kim2016deeply} proposes a deeply-recursive convolutional network (DRCN) with recursive-supervision and skip-connection to ease the training.

%

\subsection{Joint image SR}

Compared with single image SR, joint image SR attempts to leverage an additional guidance image to aid the SR process for the target image, by transferring structural information of the guidance image to the target image. 

The bilateral filter\cite{tomasi1998bilateral} is a widely used translation-variant edge-preserving filter that outputs a pixel as a weighted average of neighboring pixels. The weights are computed by a spatial filter kernel and a range filter kernel evaluated on the data values themselves. It smoothes the image while preserving edges. The joint bilateral upsampling\cite{kopf2007joint} generalizes the bilateral filter by computing the weights with respect to another guidance image rather than the input image. In particular, it applies the range filter kernel to a HR guidance image, expecting to incorporate the high frequencies of the guidance image into the LR target image. However, it has been noticed that joint bilateral image filtering may introduce gradient reversal artifacts as it does not preserve gradient information\cite{he2010guided}. Later, guided image filtering\cite{he2010guided} was proposed to overcome this limitation. Directly transferring guidance gradients can also result in notable appearance change\cite{shen2015multispectral}. To address this problem, \cite{shen2015multispectral} proposes a framework that optimizes a novel scale map to capture the nature of structure discrepancy between images. However, as the construction of these filters considers unilaterally the static guidance image, \cite{shen2015multispectral} suffers from the inconsistency of the local structures in the guidance and target images, and may therefore transfer incorrect structure details to the target images. The study in~\cite{ham2017robust} proposes robust guided image filtering, referred to as static/dynamic (SD) filtering, which jointly leverages static guidance image and dynamic target image to iteratively refine the target image. These techniques use hand-crafted objective functions that may not reflect natural image priors well. 

Recent work\cite{li2016deep} proposes a Convolutional Neural Network (CNN) based joint image filtering approach. This approach considers the structures of both input and guidance images, but requires numerous labelled images and intensive computing resources to train the deep model for each task.

Our joint image SR based on coupled dictionary learning falls into the learning-based category. Therefore, the priors used in our approach are learned from a training dataset rather than being hand-crafted and thus adapt to the target modality and guidance modality. 

\vspace{-0.3cm}

\subsection{Other multimodal image processing approaches based on sparse representations induced by a set of dictionaries}
A number of multimodal image processing approaches based on sparse representations induced by a set of dictionaries have also been proposed in the literature\cite{wang2012semi,zhuang2013supervised,liu2014semi,jing2015super,dao2016collaborative,bahrampour2016multimodal,deligiannis2017multi,deligiannis2016Xray}. However, these approaches differ from our proposed approach in a number of ways.
For example, semi-couple dictionary learning~\cite{wang2012semi}, supervised coupled dictionary learning\cite{zhuang2013supervised}, semi-supervised coupled dictionary learning\cite{liu2014semi}, and semi-coupled low-rank discriminant dictionary learning\cite{jing2015super} assume the existence of a function that maps the sparse representation of one modality to the sparse representation of another modality. In contrast, our approach does not constrain the model to require the existence of such a mapping function; instead, both similarities and disparities between different modalities are considered under the sparse representation invariance assumption. 
In turn, Dao et al.\cite{dao2016collaborative} propose a joint/collaborative sparse representation framework for multi-sensor classification. However, the dictionaries used in their work are directly constructed from training data samples and involve no dictionary learning. In comparison, the dictionaries in our work are learned from training data.
Moreover, Bahrampour et al.\cite{bahrampour2016multimodal} propose a multimodal task-driven dictionary learning algorithm under the group sparsity prior to enforce collaborations among multiple homogeneous/heterogeneous sources of information. One common feature of these works is that the sparse representations for different modalities are required to share the same support, usually induced by group sparsity, and their values are related by a mapping function.
In comparison, our model takes into account both the similarity and the discrepancy of different modalities via considering their common and unique sparse representations. This makes our approach more robust to inconsistencies between the guidance and the target images, as both of them are considered during the estimation of the target HR image, instead of unilaterally transferring the structure details from the guidance image. 
%
Our data model used in our multimodal image SR approach is inspired from the data model proposed in~\cite{deligiannis2017multi,deligiannis2016Xray} used for multimodal image separation. However, the generalization of the approach from multimodal image separation to multimodal image SR entails a number of innovations including: (1) unique dictionaries are introduced for the side information because we consider that the side information also contains its own unique features; (2) both our coupled dictionary learning and coupled SR algorithms are different from~\cite{deligiannis2017multi,deligiannis2016Xray}. Overall, practical experiments demonstrate that the proposed multimodal image SR approach outperforms the state-of-the-art in various scenarios.
\begin{figure*}[ht]
	\centering	
	{
		\centering
		\includegraphics[width= 13cm]{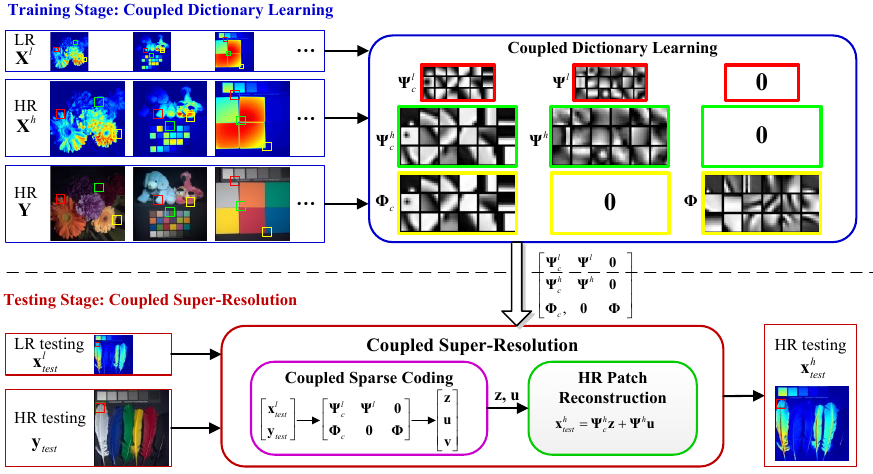}  
	}
	%
	%
	
	\caption{Proposed multimodal image super-resolution approach encompassing both a training stage and a testing stage. $\mathbf{X}$ (or $\mathbf{x}$) and $\mathbf{Y}$ (or $\mathbf{y}$) represent the target and guidance modalities, respectively. }
	\label{Fig:Diagram}
\end{figure*}

\section{Multimodal Image SR via Joint Sparse Representations Induced by Coupled Dictionaries}
\label{sec:SIMIS}
We now introduce our SR approach. In particular, we describe the data model that couples different image modalities and also the joint image SR framework that encompasses both a coupled dictionary learning phase and a coupled super-resolution phase, see also Figure~\ref{Fig:Diagram}).
%


\subsection{Multimodal Data Model}


\mypar{Basic Data Model}
It is commonly observed that images of different modalities contain similarities as well as disparities. These characteristics can be effectively modelled in a sparse feature space so that different modalities can be related together via their sparse representations with respect to a group of coupled dictionaries. We first introduce a basic data model that captures the relationships -- including similarities and disparities -- between two different image modalities. In particular, we propose to use joint sparse representations to express a pair of registered, vectorized image patches $\mathbf{x} \in \mathbb{R}^{N_x} $ and $\mathbf{y} \in \mathbb{R}^{N_y}$ associated with different modalities as follows:
%
\begin{equation} \label{Eq:MultimodalDataModel}
\begin{split}
\mathbf{x} 
&= \boldsymbol{\Psi}_{c} \, \mathbf{\mathbf{z}} + \boldsymbol{\Psi} \, \mathbf{u} \,,
\\
\mathbf{y} 
&= \boldsymbol{\Phi}_{c} \, \mathbf{\mathbf{z}} + \boldsymbol{\Phi} \, \mathbf{v} \,,
\end{split}
\end{equation}
where $\mathbf{z} \in \mathbb{R}^{K_c}$ is a sparse representation that is common to both modalities, $\mathbf{u} \in \mathbb{R}^{K_u}$ is a sparse representation specific to modality $\mathbf{x}$, while $\mathbf{v} \in \mathbb{R}^{K_v}$ is a sparse representation specific to modality $\mathbf{y}$. In turn, $\boldsymbol{\Psi}_{c} \in \mathbb{R}^{N_x \times K_c}$ and $\boldsymbol{\Phi}_{c} \in \mathbb{R}^{N_y \times K_c}$ are a pair of dictionaries associated with the common sparse representation $\mathbf{z}$, whereas $\boldsymbol{\Psi} \in \mathbb{R}^{N_x \times K_u}$ and $\boldsymbol{\Phi} \in \mathbb{R}^{N_y \times K_v}$ are dictionaries associated with the specific sparse representations $\mathbf{u}$ and $\mathbf{v}$, respectively. (For simplicity, we take $N = N_x = N_y$, $K = K_c = K_u = K_v$ hereafter.)

\mypar{SR Data Model}
We now transform the basic data model in~\eqref{Eq:MultimodalDataModel} into the SR data model that underlies our proposed super-resolution process. This model is based on two main assumptions:

1. First, we assume -- as in~\eqref{Eq:MultimodalDataModel} -- that similarities and disparities between the LR and HR versions of the patches of the different image modalities can be captured using sparse representations.

2. Second, we also assume -- as in~\cite{yang2008image,yang2010image,zeyde2010single} -- that the LR and HR versions of a patch of a certain image modality share the same sparse representation, albeit not the same dictionary.

In particular, we express the LR image patch $\mathbf{x}^{l} \in \mathbb{R}^{M}$ and HR image patch $\mathbf{x}^{h} \in \mathbb{R}^{N} $ of a certain image modality, and another HR registered patch of another corresponding image modality $\mathbf{y} \in \mathbb{R}^{N}$ as follows:\footnote{
Our model assumes identical common sparse representations so that each pair of common atoms is adjusted automatically to satisfy this assumption. In addition, we also take into account the discrepancy of different modalities via considering their unique sparse representations. This differs from the models used in~\cite{wang2012semi,zhuang2013supervised,liu2014semi,jing2015super,dao2016collaborative,bahrampour2016multimodal,deligiannis2017multi,deligiannis2016Xray}, some of which 
assume that the sparse representations for different modalities share the same support and some assume that they share identical sparse representations without consideration to the discrepancy.
}
%
%
\begin{align}
\mathbf{x}^{h} &= \boldsymbol{\Psi}_{c}^{h} \, \mathbf{z} + \boldsymbol{\Psi}^{h} \, \mathbf{u} \,,
\label{Eq:SparseModelX}
\\
\mathbf{x}^{l} 
&= \boldsymbol{\Psi}_{c}^{l} \, \mathbf{z} + \boldsymbol{\Psi}^{l} \, \mathbf{u} \,,
\label{Eq:SparseModelX_low}
\\
\mathbf{y} &= \boldsymbol{\Phi}_{c} \, \mathbf{z} + \boldsymbol{\Phi} \, \mathbf{v} \,,
\label{Eq:SparseModelY}
\end{align}
where, as in the basic data model~\eqref{Eq:MultimodalDataModel}, $\mathbf{z} \in \mathbb{R}^{K}$ is the common sparse representation shared by both modalities, $\mathbf{u} \in \mathbb{R}^{K}$ is the unique sparse representation specific to modality $\mathbf{x}$ while $\mathbf{v} \in \mathbb{R}^{K}$ is the unique sparse representation specific to modality $\mathbf{y}$. In turn, $\boldsymbol{\Psi}_{c}^{h} \in \mathbb{R}^{N \times K}$, $\boldsymbol{\Psi}_{c}^{l} \in \mathbb{R}^{M \times K}$ and $\boldsymbol{\Phi}_{c} \in \mathbb{R}^{N \times K}$ are the dictionaries associated with the common sparse representation $\mathbf{z}$, whereas $\boldsymbol{\Psi}^{h} \in \mathbb{R}^{N \times K}$, $\boldsymbol{\Psi}^{l} \in \mathbb{R}^{M \times K}$ and $\boldsymbol{\Phi} \in \mathbb{R}^{N \times K}$ are dictionaries associated with the specific sparse representations $\mathbf{u}$ and $\mathbf{v}$, respectively.
Note that the sparse vectors $\mathbf{z}$ and $\mathbf{u}$ capture the relationship between the LR and HR patches of the same modality in \eqref{Eq:SparseModelX} and \eqref{Eq:SparseModelX_low}. Moreover, the common sparse vector $\mathbf{z}$ connects the various patches of the two different modalities in \eqref{Eq:SparseModelX} - \eqref{Eq:SparseModelY}. The disparities between modalities $\mathbf{x}$ and $\mathbf{y}$ are distinguished by the sparse vectors $\mathbf{u}$ and $\mathbf{v}$. 
Overall, this data model allows each pair of patches to be non-linearly transformed to a sparse domain with respect to a group of coupled dictionaries in order to obtain sparse representations that characterize the similarities and disparities between different modalities. Note also that our data model reduces to the data model in \cite{yang2010image,yang2012coupled,zeyde2010single} -- applicable to single modality image super-resolution -- provided that the side information $\mathbf{y}$ is neglected.


By capitalizing on this model, we propose in the sequel a novel joint image SR scheme that consists of two stages: (1) a training stage referred to as coupled dictionary learning (CDL) and (2) a testing stage referred to as coupled image super-resolution (CSR) (see Figure~\ref{Fig:Diagram}). In the training stage, we learn the dictionaries in \eqref{Eq:SparseModelX} - \eqref{Eq:SparseModelY} from a set of training image patches to couple different data modalities together. Then, in the testing stage, we use the learned dictionaries to find the representations of the LR testing patch and corresponding HR guidance patch, according to \eqref{Eq:SparseModelX_low} and \eqref{Eq:SparseModelY}. These sparse representations are then used to reconstruct the desired HR target image patch via \eqref{Eq:SparseModelX}.


\vspace{-0.4cm}

\subsection{Coupled Dictionary Learning (CDL)}
\label{ssec:CDL}

\begin{algorithm*}[t] 
	\caption{Coupled Dictionary Learning}
	\label{Alg:CK-SVD}
	\begin{multicols}{2}  
	\begin{algorithmic}[1]
		%
		\renewcommand{\algorithmicrequire}{\textbf{Input:}}
		\renewcommand{\algorithmicensure}{\textbf{Output:}}
		\REQUIRE 
		Training data matrices $\mathbf{X}^l$, $\mathbf{X}^h$ and $\mathbf{Y}$.
		
		\ENSURE 
		Dictionary pairs $[\boldsymbol{\Psi}_{c}^l,\boldsymbol{\Psi}^l]$, $[\boldsymbol{\Psi}_{c}^h,\boldsymbol{\Psi}^h]$ and $[\boldsymbol{\Phi}_{c},\boldsymbol{\Phi}]$.
		%
		%
		\renewcommand{\algorithmicrequire}{\textbf{Initialization:}}
		\REQUIRE
		Initialize dictionary atoms with randomly selected patches. Set the training iterations $OutIter$ and $InIter$, sparsity constraint $s$ and residual constraint $\epsilon$.
		
		\renewcommand{\algorithmicrequire}{\textbf{Optimization:}}
		\REQUIRE
		
		\STATE
		\textbf{Step 1 -- LR Dictionary learning:} 
		\FOR{$p = 1$ to $OutIter$} 
		\label{AlgOpt:OutIterStart}
		\FOR{$q = 1$ to $InIter$}
		\STATE  \label{AlgOpt:GlobalSC1}
		\label{Code:SRO:GlobalSparseCoding}
		\textbf{Global Sparse Coding}. 
		Fix all the dictionaries, then solve~\eqref{Eq:CDL_SparseCoding} to update sparse representations $\mathbf{Z}$, $\mathbf{U}$ and $\mathbf{V}$ by performing OMP on each training example.
		
		\STATE 
		\label{Alg:OMPstart}
		Initialize the active set $\Gamma=\emptyset$ and $[\mathbf{z}_{i}^T; \mathbf{u}_{i}^T; \mathbf{v}_{i}^T] \leftarrow \mathbf{0}$.
		\WHILE{$|\Gamma| < s_c$ or or residual $>\epsilon$} 
		\STATE
		select a new coordinate $\hat{k}$ that leads to the smallest residual and, then update the active set and the sparse representations:
		{\small 
		\begin{align*}
		&
		(\hat{k},\hat{\boldsymbol{\alpha}}) \in 
		\underset{k \in \Gamma^c, \boldsymbol{\alpha} \in \mathbb{R}^{|\Gamma|+1}}{\arg \min } \;
		\left\| 
		\begin{bmatrix} 
		\mathbf{x}_{i}^l  \\ 
		\mathbf{y}_{i} 
		\end{bmatrix} 
		- 	
		\begin{bmatrix} 
		\boldsymbol{\Psi}_c^l \; \boldsymbol{\Psi}^l \; \mathbf{0} \\ 
		\boldsymbol{\Phi}_c  \;\;  \mathbf{0} \; \boldsymbol{\Phi}_c
		\end{bmatrix}_{\Gamma \cup \{k\}}
		\boldsymbol{\alpha}
		\right\|_2^2
		\\
		&
		\Gamma \leftarrow \Gamma \cup \{\hat{k}\}; \, 
		[\mathbf{z}_{i}^T; \mathbf{u}_{i}^T; \mathbf{v}_{i}^T]_{\Gamma} \leftarrow \hat{\boldsymbol{\alpha}}; \, 
		[\mathbf{z}_{i}^T; \mathbf{u}_{i}^T; \mathbf{v}_{i}^T]_{\Gamma^c} \leftarrow \mathbf{0}
		\end{align*}
		}
		\ENDWHILE
		\label{Alg:OMPend}

		\STATE \label{AlgOpt:LocalDictUpdate1}
		\textbf{Local Common Dictionary Update}. 
		Fix $\boldsymbol{\Psi}^l$, $\boldsymbol{\Phi}$, and only update $\boldsymbol{\Psi}^l_{c}$ and $\boldsymbol{\Phi}_{c}$ by solving~\eqref{Eq:PartialDictionaryUpdate_Com}.
		Specifically, for each atom pair 
		$\begin{bmatrix} 
		\boldsymbol{\psi}_{ck}^l \\ \boldsymbol{\phi}_{ck} 			 
		\end{bmatrix}$ of
		$\begin{bmatrix} 
			\boldsymbol{\Psi}_{c}^l \\ \boldsymbol{\Phi}_{c} 			 
		\end{bmatrix}$,
		denote by $\mathbf{z}^k$ the $k$-th row vector in $\mathbf{Z}$, and $\Omega_k = \{ i | 1 \leq i \leq T, \mathbf{z}^k(i) \neq 0 \}$ the index set of those training samples that use $k$-th atom pair. Then, compute the representation residual
			\begin{equation*}
 				\mathbf{E}_k =  
 				\left(
 				\begin{bmatrix} 
 				\mathbf{X}^l - \boldsymbol{\Psi}^l \mathbf{U}
 				\\ 
 				\mathbf{Y} - \boldsymbol{\Phi} \mathbf{V}			 \end{bmatrix}
 				- 
 				\begin{bmatrix} 
 				\boldsymbol{\Psi}_{c}^l \\ \boldsymbol{\Phi}_{c} 			 
 				\end{bmatrix}
 				\mathbf{Z}
 				+ 
 				\begin{bmatrix} 
 				\boldsymbol{\psi}_{ck}^l \\ \boldsymbol{\phi}_{ck} 			 
 				\end{bmatrix} \mathbf{z}^k
 				\right)_{(:,\Omega_k)} 
			\end{equation*}
			Apply SVD on $\mathbf{E}_k = \mathbf{P} \mathbf{\Sigma} \mathbf{Q}^\text{T}$ and choose the first column of $ \mathbf{P} $ as the updated atom pair
			$\begin{bmatrix} 
			\boldsymbol{\psi}_{ck}^l \\ \boldsymbol{\phi}_{ck} 			 
			\end{bmatrix}$. 
		%
		\ENDFOR
		\FOR{$q = 1$ to $InIter$}
		\STATE \label{AlgOpt:GlobalSC2}
		\textbf{Global Sparse Coding}. The same as step \ref{Code:SRO:GlobalSparseCoding}.
		\STATE \label{AlgOpt:LocalDictUpdate2}
		\textbf{Local Unique Dictionary Update}. 
		Fix $\boldsymbol{\Psi}^l_{c}$, $\boldsymbol{\Phi}_{c}$, and only update $\boldsymbol{\Psi}^l$ and $\boldsymbol{\Phi}$ by solving ~\eqref{Eq:PartialDictionaryUpdate_Psi} and \eqref{Eq:PartialDictionaryUpdate_Phi}.
		For each atom  
		$\boldsymbol{\psi}_{k}^l$ of $\boldsymbol{\Psi}^l$,
		%
		denote by $\mathbf{u}^k$ the $k$-th row vector in $\mathbf{U}$, and $\Omega_k = \{ i | 1 \leq i \leq T, \mathbf{u}^k(i) \neq 0 \}$. Then, compute the representation residual
		\begin{equation*}
		\mathbf{E}_k =  
		\left(
		\begin{bmatrix} 
		\mathbf{X}^l - \boldsymbol{\Psi}^l_{c} \mathbf{Z}			 
		\end{bmatrix}
		- 
		\boldsymbol{\Psi}^l \mathbf{U} 
		+ 
		\boldsymbol{\psi}_{k}^l \mathbf{u}^k 
		\right)_{(:,\Omega_k)} 
		\end{equation*}
		Apply SVD on $\mathbf{E}_k = \mathbf{P} \mathbf{\Sigma} \mathbf{Q}^\text{T}$ and choose the first column of $ \mathbf{P} $ as the updated atom $\boldsymbol{\psi}_{k}^l$. Each atom $\boldsymbol{\phi}_{k}$ of $\boldsymbol{\Phi}$ is updated with $\Omega_k = \{ i | 1 \leq i \leq T, \mathbf{v}^k(i) \neq 0 \}$ and
		$
		\mathbf{E}_k =  
		\left(
		\begin{bmatrix} 
		\mathbf{Y} - \boldsymbol{\Phi}_{c} \mathbf{Z}			 
		\end{bmatrix}
		- 
		\boldsymbol{\Phi} \mathbf{V} 
		+ 
		\boldsymbol{\phi}_{k} \mathbf{v}^k 
		\right)_{(:,\Omega_k)} 
		$
		in a similar manner.	
		\ENDFOR
		
		\ENDFOR
		\label{AlgOpt:OutIterEnd}

		\STATE 
		\textbf{Step 2 -- HR Dictionary learning:} 
		\STATE
		Construct $[\boldsymbol{\Psi}_{c}^h,\boldsymbol{\Psi}^h]$ as in \eqref{Eq:CoupledDL_DL3}.
		
		\STATE
		Return dictionaries.
	\end{algorithmic}
	\end{multicols}
\end{algorithm*}

We assume that we have access to $T$ registered patches of LR, HR and guidance images for learning our data model in \eqref{Eq:SparseModelX} - \eqref{Eq:SparseModelY}. In particular, let $\mathbf{x}^l_i$, $\mathbf{x}^h_i$ and $\mathbf{y}_i$ ($i=1 \ldots T$) denote the registered patches corresponding to the LR, HR, and the guidance training image patches, and let $\mathbf{z}_i$, $\mathbf{u}_i$ and $\mathbf{v}_i$ ($i=1 \ldots T$) denote their sparse representations. Our coupled dictionary learning problem can now be posed as follows:
\begin{equation} \label{Eq:CoupledDL}
\begin{array}{cl}
\underset{ 
	\begin{subarray}{c}
	\left\{ \boldsymbol{\Psi}_{c}^l, \boldsymbol{\Psi}^l, \boldsymbol{\Psi}_{c}^h, \right. \\
	\left. \boldsymbol{\Psi}^h, \boldsymbol{\Phi}_{c}, \boldsymbol{\Phi} \right\}  \\
	\{ \mathbf{Z}, \mathbf{U}, \mathbf{V} \}
	\end{subarray}}
{ \text{minimize}}
& 
\left\|
\begin{bmatrix} 
\mathbf{X}^l \\ \mathbf{X}^h \\ \mathbf{Y} 
\end{bmatrix}
-
\begin{bmatrix}
\boldsymbol{\Psi}_{c}^l & \boldsymbol{\Psi}^l & \mathbf{0} \\
\boldsymbol{\Psi}_{c}^h & \boldsymbol{\Psi}^h & \mathbf{0} \\
\boldsymbol{\Phi}_{c} & \mathbf{0} & \boldsymbol{\Phi} \\
\end{bmatrix}
\begin{bmatrix}
\mathbf{Z} \\
\mathbf{U} \\
\mathbf{V} \\
\end{bmatrix}
\right\|_F^2
\\
\text{subject to}
& 
\|\mathbf{z}_i \|_0 
+ \|\mathbf{u}_i \|_0
+ \|\mathbf{v}_i \|_0 \leq s, \; \forall i,
\end{array}
\end{equation}
where $\mathbf{X}^l = \left[\mathbf{x}^l_1,...,\mathbf{x}^l_T\right]$ $\in \mathbb{R}^{M \times T}$, $\mathbf{X}^h = \left[\mathbf{x}^h_1,...,\mathbf{x}^h_T\right]$ $\in \mathbb{R}^{N \times T}$ and $\mathbf{Y} = \left[\mathbf{y}_1,...,\mathbf{y}_T \right]$ $\in \mathbb{R}^{N \times T}$, $\mathbf{Z} = \left[\mathbf{z}_1, ..., \mathbf{z}_T \right]$ $\in \mathbb{R}^{K \times T}$, $\mathbf{U} = \left[\mathbf{u}_1, ..., \mathbf{u}_T \right]$ $\in \mathbb{R}^{K \times T}$ and $\mathbf{V} = \left[\mathbf{v}_1, ..., \mathbf{v}_T \right]$ $\in \mathbb{R}^{K \times T}$, and $\| \cdot \|_F $ and $\| \cdot \|_0 $ denote the Frobenius norm and $\ell_0$ pseudo-norm, respectively.

Note that -- akin to other dictionary learning formulations\cite{aharon2006img}  -- the objective in the optimization problem~\eqref{Eq:CoupledDL} encourages the data representation to approximate the data, and the constraint in~\eqref{Eq:CoupledDL} encourages the data representation to be sparse (i.e. the overall sparsity of the data representations is constrained to be less than or equal to $s$)\footnote{
Note that, we could also use alternative sparsity constraints, such as (a) $\|\mathbf{z}_i \|_0 + \|\mathbf{u}_i \|_0 \leq s_x, \|\mathbf{z}_i \|_0 + \|\mathbf{v}_i \|_0 \leq s_y$, (b) $\|\mathbf{z}_i \|_0 \leq s_z, \|\mathbf{u}_i \|_0 \leq s_u, \|\mathbf{v}_i \|_0 \leq s_v.$ Empirical studies suggest that these constraints lead to similar performance. We prefer the constraint in~\eqref{Eq:CoupledDL} since it makes the formulation concise, with fewer parameters for tuning.
%
%
}.

We address the coupled dictionary learning problem \eqref{Eq:CoupledDL} in two steps: LR Dictionary learning and HR Dictionary learning. In the first step (LR Dictionary learning), the algorithm uses LR patches $\mathbf{X}^l$ and side information $\mathbf{Y}$ to learn the two pairs of dictionaries $[\boldsymbol{\Psi}_c^l, \boldsymbol{\Psi}^l]$ and $[\boldsymbol{\Phi}_c, \boldsymbol{\Phi}]$ and the sparse codes $\mathbf{Z}$, $\mathbf{U}$, $\mathbf{V}$, via solving a non-convex optimization problem. In the second step (HR Dictionary learning), the algorithm uses HR patches $\mathbf{X}^h$ and the sparse codes $\mathbf{U}$, $\mathbf{V}$ to learn the HR dictionaries $[\boldsymbol{\Psi}_c^h, \boldsymbol{\Psi}^h]$.\footnote{
	The motivation of this two-step training strategy is that the sparse codes $\mathbf{Z}$ and $\mathbf{U}$ should be obtained only from $\mathbf{X}^l$ and $\mathbf{Y}$ in both training and testing stages without involving $\mathbf{X}^h$, since the HR target patches $\mathbf{X}^h$ are available only in the training stage and not in testing stage. Similar strategies are also adopted by other works~\cite{zeyde2010single} and the empirical results suggest better performance.
}
Algorithm~\ref{Alg:CK-SVD} shows how we adapt K-SVD [46] accordingly.

\subsubsection{Step 1 -- LR Dictionary learning}
In the first step, we learn the dictionary pairs $[\boldsymbol{\Psi}_c^l, \boldsymbol{\Psi}^l]$, $[\boldsymbol{\Phi}_c, \boldsymbol{\Phi}]$ and the sparse codes $\mathbf{Z}$, $\mathbf{U}$, $\mathbf{V}$ from $\mathbf{X}^l$ and $\mathbf{Y}$ by solving the following optimization problem:
%
\begin{equation} \label{Eq:CoupledDL_DL1}
\begin{array}{cl}
\underset{ 
	\begin{subarray}{c}
	\left\{ \boldsymbol{\Psi}_{c}^l, \boldsymbol{\Psi}^l, \boldsymbol{\Phi}_{c}, \boldsymbol{\Phi} \right\}  \\
	\{ \mathbf{Z}, \mathbf{U}, \mathbf{V} \}
	\end{subarray}}
{ \text{minimize}}
& 
\left\|
\begin{bmatrix} 
\mathbf{X}^l \\ \mathbf{Y} 
\end{bmatrix}
-
\begin{bmatrix}
\boldsymbol{\Psi}_{c}^l & \boldsymbol{\Psi}^l & \mathbf{0} \\
\boldsymbol{\Phi}_{c} & \mathbf{0} & \boldsymbol{\Phi} \\
\end{bmatrix}
\begin{bmatrix}
\mathbf{Z} \\
\mathbf{U} \\
\mathbf{V} \\
\end{bmatrix}
\right\|_F^2
\\
\text{subject to}
& 
\|\mathbf{z}_i \|_0 
+ \|\mathbf{u}_i \|_0
+ \|\mathbf{v}_i \|_0 \leq s, \; \forall i.
\end{array}
\end{equation}

\noindent
In order to handle this non-convex optimization problem, we adopt an alternating optimization approach that performs sparse coding and dictionary update alternatively. 

During the sparse coding stage, we first fix the global dictionaries and obtain the sparse representations by solving:
\begin{equation} \label{Eq:CDL_SparseCoding}
\begin{array}{cl}
\underset{\mathbf{Z}, \mathbf{U}, \mathbf{V}}{ \text{min}}
& 
\left\|
\begin{bmatrix} \mathbf{X}^l \\ \mathbf{Y} \end{bmatrix}
-
\begin{bmatrix}
\boldsymbol{\Psi}^l_{c} & \boldsymbol{\Psi}^l & \mathbf{0} \\
\boldsymbol{\Phi}_{c} & \mathbf{0} & \boldsymbol{\Phi} \\
\end{bmatrix}
\begin{bmatrix}
\mathbf{Z} \\
\mathbf{U} \\
\mathbf{V} \\
\end{bmatrix}
\right\|_F^2
\\
\text{s.t.}
& 
\|\mathbf{z}_i \|_0 
+ \|\mathbf{u}_i \|_0
+ \|\mathbf{v}_i \|_0 \leq s, \; \forall i.
\end{array}
\end{equation}
\noindent
This problem -- which we call global sparse coding because it updates all the sparse representations $\mathbf{Z}$, $\mathbf{U}$ and $\mathbf{V}$ -- is solved using the orthogonal matching pursuit (OMP) algorithm \cite{tropp2007signal}.\footnote{An additional error threshold parameter $\epsilon$ is used to deal with noisy images. This parameter defines whether or not one should stop the OMP loop depending on the residual of the objective. See Algorithm~\ref{Alg:CK-SVD}.
}

During the dictionary updating stage, we fix the sparse codes and update the global dictionaries via solving:
\begin{equation} \label{Eq:CDL_DictUpdate}
\begin{array}{cl}
\underset{ 
	\begin{subarray}{c}
	 \boldsymbol{\Psi}_{c}^l, \boldsymbol{\Psi}^l, \boldsymbol{\Phi}_{c}, \boldsymbol{\Phi} 
	\end{subarray}}
{ \text{minimize}}
& 
\left\|
\begin{bmatrix} 
\mathbf{X}^l \\ \mathbf{Y} 
\end{bmatrix}
-
\begin{bmatrix}
\boldsymbol{\Psi}_{c}^l & \boldsymbol{\Psi}^l & \mathbf{0} \\
\boldsymbol{\Phi}_{c} & \mathbf{0} & \boldsymbol{\Phi} \\
\end{bmatrix}
\begin{bmatrix}
\mathbf{Z} \\
\mathbf{U} \\
\mathbf{V} \\
\end{bmatrix}
\right\|_F^2 \,.
\end{array} 
\end{equation}

\noindent
To this end, we adapt the K-SVD\cite{aharon2006img} algorithm for our coupled dictionary learning case. The key idea is to update common dictionaries simultaneously while updating unique dictionaries individually.\footnote{
	Owing to the SVD operation in the dictionary update, atoms from the common dictionary pair [$\boldsymbol{\Psi}^l_{c}$; $\boldsymbol{\Phi}_{c}$] and the unique dictionaries $\boldsymbol{\Psi}^l$ and $\boldsymbol{\Phi}$ have unit $\ell_2$ norm automatically.
}
%
Specifically, we further decompose Problem~\eqref{Eq:CoupledDL_DL1} into the following convex sub-problems~\eqref{Eq:PartialDictionaryUpdate_Com} - \eqref{Eq:PartialDictionaryUpdate_Phi}, so that we can sequentially learn the common dictionaries and the unique dictionaries. That is, we fix the unique dictionaries $\boldsymbol{\Psi}^l$, $\boldsymbol{\Phi}$ and only update the common dictionaries $\boldsymbol{\Psi}^l_{c}$ and $\boldsymbol{\Phi}_{c}$ by solving
%
%
\begin{equation} \label{Eq:PartialDictionaryUpdate_Com}
\begin{array}{cl}
\underset{
	\begin{subarray}{c}
	\boldsymbol{\Psi}^l_{c}, \boldsymbol{\Phi}_{c}	\\
	\end{subarray}}{ \text{min}}
&
\left\|
\begin{bmatrix} 
\mathbf{X}^l - \boldsymbol{\Psi}^l \mathbf{U} \\
\mathbf{Y} - \boldsymbol{\Phi} \mathbf{V}
\end{bmatrix}
-
\begin{bmatrix}
\boldsymbol{\Psi}^l_{c}  \\
\boldsymbol{\Phi}_{c}  \\
\end{bmatrix}
\mathbf{Z}
\right\|_F^2 \,.
\end{array}
\end{equation}
The algorithm alternates between global sparse coding~\eqref{Eq:CDL_SparseCoding} and local common dictionary update~\eqref{Eq:PartialDictionaryUpdate_Com} for a few iterations until the procedure converges. Next, we fix the already learned common dictionaries and train the unique dictionaries by alternating between global sparse coding~\eqref{Eq:CDL_SparseCoding} and following two unique dictionary update operations:
%
%
\begin{equation} \label{Eq:PartialDictionaryUpdate_Psi}
\begin{array}{cl}
\underset{\boldsymbol{\Psi}^l}{ \text{min}}
&
\left\|
\left( 
\mathbf{X}^l - \boldsymbol{\Psi}^l_{c} \mathbf{Z} 
\right)
- \boldsymbol{\Psi}^l \mathbf{U}
\right\|_F^2 \,.
\end{array}
\end{equation}
\begin{equation} \label{Eq:PartialDictionaryUpdate_Phi}
\begin{array}{cl}
\underset{\boldsymbol{\Phi}}{ \text{min}}
&
\left\|
\left(
\mathbf{Y}
-		
\boldsymbol{\Phi}_{c} \mathbf{Z}
\right)
-
\boldsymbol{\Phi} \mathbf{V}
\right\|_F^2 \,.
\end{array}
\end{equation}	


\subsubsection{Step 2 -- HR Dictionary learning}
In the second step,  once the dictionary pairs $[\boldsymbol{\Psi}_c^l, \boldsymbol{\Psi}^l]$ and $[\boldsymbol{\Phi}_c, \boldsymbol{\Phi}]$ are learned from $\mathbf{X}^l$ and $\mathbf{Y}$, we construct the HR dictionaries $[\boldsymbol{\Psi}_c^h, \boldsymbol{\Psi}^h]$ based on $\mathbf{X}^h$ and sparse codes $\mathbf{Z}$ and $\mathbf{U}$ by solving the optimization problem: 
%
%
%
\begin{equation} \label{Eq:CoupledDL_DL2}
\begin{array}{cl}
\underset{ \boldsymbol{\Psi}_{c}^h, \boldsymbol{\Psi}^h }
{ \text{min}}
& \! \! \!
\left\|
\mathbf{X}^h
-
\boldsymbol{\Psi}_{c}^h \mathbf{Z} - \boldsymbol{\Psi}^h \mathbf{U} 
\right\|_F^2
+
\lambda
\left\|
\begin{bmatrix}
\boldsymbol{\Psi}_{c}^h & \boldsymbol{\Psi}^h \\
\end{bmatrix}
\right\|_F^2
\end{array}
\end{equation}
where the second term serves as a regularizer that makes the solution more stable\footnote{In order to guarantee the fidelity of the sparse approximation to the HR training datasets, the atoms in the HR dictionaries are not constrained to be unit $\ell_2$ norm, as in~\cite{zeyde2010single} and~\cite{wang2010resolution}. However, when there are zeros or near-zeros rows in the sparse codes $\mathbf{Z}$ or $\mathbf{U}$, the matrix inverse operation during the computation of the closed form solution will give extremely large value for corresponding atoms. Therefore, in order to make the solution more stable, a Frobenius norm is added to regularize Problem \eqref{Eq:CoupledDL_DL2}.}.
This optimization problem -- which exploits the conventional sparse representation invariance assumption that HR image patches $\mathbf{X}^h$ share the same sparse codes with the corresponding LR version $\mathbf{X}^l$ -- admits the closed form solution
\begin{equation} \label{Eq:CoupledDL_DL3}
\begin{bmatrix}
\boldsymbol{\Psi}_{c}^h & \boldsymbol{\Psi}^h \\
\end{bmatrix}
=
\mathbf{X}^h \Gamma^T (\Gamma \Gamma^T + \lambda \mathbf{I})^{-1} \,,
\text{where, } 
\Gamma = \begin{bmatrix} \mathbf{Z} \\ \mathbf{U} \end{bmatrix}
\end{equation}

Similar to conventional dictionary learning, our CDL algorithm cannot guarantee the convergence to a global optimum due to the non-convexity nature of Problem \eqref{Eq:CoupledDL}. However, CDL is convex with respect to the dictionaries when the sparse codes are fixed or vice versa. This property ensures that dictionary algorithms usually converge to a local optimum that leads to good SR performance. This is also confirmed by experiments on both real and synthetic data, presented in Section \ref{sec:Experiments}.


\subsection{Coupled Super Resolution (CSR)}
Given the learned coupled dictionaries associated with the model in \eqref{Eq:SparseModelX} - \eqref{Eq:SparseModelY}, we now assume that we have access to a LR testing image and a corresponding registered HR guidance image as side information. We extract (overlapping) image patch pairs from these two modalities. In particular, let $\mathbf{x}^{l}_{test} \in \mathbb{R}^M$ denote a LR testing image patch and let $\mathbf{y}^{h}_{test} \in \mathbb{R}^N$ denote the corresponding HR guidance image patch. We can now pose a coupled super-resolution problem that involves two steps.

\subsubsection{Step 1 -- Coupled Sparse Coding}
First, we solve the optimization problem

%
%
\begin{equation} \label{Eq:SR}
\begin{array}{cl}
\underset{\mathbf{z},\mathbf{u},\mathbf{v}}{\text{min}} 
& 
\left\|
\begin{bmatrix} 
\mathbf{x}^{l}_{test} \\ \mathbf{y}_{test}
\end{bmatrix}
-
\begin{bmatrix}
\boldsymbol{\Psi}_{c}^l & \boldsymbol{\Psi}^l & \mathbf{0} \\
\boldsymbol{\Phi}_{c} & \mathbf{0} & \boldsymbol{\Phi} \\
\end{bmatrix}
\begin{bmatrix}
\mathbf{z} \\
\mathbf{u} \\
\mathbf{v} \\
\end{bmatrix}
\right\|_2^2
\\
\text{s.t.}
& 
\|\mathbf{z} \|_0 
+ \|\mathbf{u} \|_0
+ \|\mathbf{v} \|_0 \leq s \,,
\end{array}
\end{equation}
%
%

\noindent
where the $\ell_2$ norm promotes the fidelity of sparse representations to the signals and the $\ell_0$ pseudo-norm promotes sparsity for the sparse codes.
Some off-the-shelf algorithms -- such as orthogonal matching pursuit (OMP) algorithm~\cite{tropp2007signal} and iterative hard-thresholding algorithm~\cite{blumensath2009iterative} -- can be applied to approximate the solution to \eqref{Eq:SR}. Compared with conventional sparse coding problems that involves only LR image patch $\mathbf{x}^{l}$, our formulations~\eqref{Eq:SR} also integrates the side information $\mathbf{y}_{test}$ into the sparse coding task. Since the increase in the amount of available information is akin to the increase of the number of measurements in a Compressive Sensing scenario~\cite{mota2017compressed,renna2016classification}, one can expect to obtain a more accurate estimate of the sparse codes.

\subsubsection{Step 2 -- HR Patch Reconstruction}
Finally, we can obtain an estimate of the HR patch of the target image $\mathbf{x}^{h}_{test}$ from the HR dictionaries $[\boldsymbol{\Psi}_c^h, \boldsymbol{\Psi}^h]$ and sparse codes $\mathbf{z}$ and $\mathbf{u}$ as follows

%
\begin{equation} \label{Eq:SR_2}
\mathbf{x}^{h}_{test} = \boldsymbol{\Psi}_{c}^{h} \mathbf{z} + \boldsymbol{\Psi}^{h} \mathbf{u} \,.
\end{equation}

\noindent
Once all the HR patches are recovered, they are integrated into a whole image by averaging on the overlapping areas. The coupled super-resolution algorithm is described in Algorithm~\ref{Alg:CSR}.

\begin{algorithm}[t] 
	\caption{Coupled Super-resolution}
	\label{Alg:CSR}
	\begin{algorithmic}[1]
		%
		\renewcommand{\algorithmicrequire}{\textbf{Input:}}
		\renewcommand{\algorithmicensure}{\textbf{Output:}}
		\REQUIRE 
		The testing patch $\mathbf{x}^{l}_{test}$ and side information $\mathbf{y}_{test}$. \\
		Learned dictionaries $[\boldsymbol{\Psi}_{c}^l,\boldsymbol{\Psi}^l]$, $[\boldsymbol{\Psi}_{c}^h,\boldsymbol{\Psi}^h]$ and $[\boldsymbol{\Phi}_{c},\boldsymbol{\Phi}]$.
		
		\ENSURE 
		High resolution estimation $\mathbf{x}^{h}_{test}$.
		\renewcommand{\algorithmicrequire}{\textbf{Operations:}}
		\REQUIRE
		\STATE 
		\textbf{Step 1 -- Coupled Sparse Coding:} 
		
		Use off-the-shelf sparse coding algorithms to solve the problem~\eqref{Eq:SR} to obtain the sparse codes
		$\mathbf{z}$, $\mathbf{u}$ and $\mathbf{v}$.
		\STATE 
		\textbf{Step 2 -- HR Patch Reconstruction:} 
		
		Reconstruct the HR patch $\mathbf{x}^{h}_{test}$ as in~\eqref{Eq:SR_2}. 
	\end{algorithmic}
\end{algorithm}

\section{Experiments}
\label{sec:Experiments}
We now present a series of experiments to validate the effectiveness of the proposed joint image SR approach in various scenarios. In subsection \ref{sec:Experiments}-A, we perform multi-spectral image super-resolution (MS-SR) aided by the corresponding RGB version of the same scene. In subsection \ref{sec:Experiments}-B, we perform near-infrared image super-resolution (NIR-SR) aided by the corresponding RGB version of the same scene. We consider situations where the training and/or testing images are contaminated by noise in subsection \ref{sec:Experiments}-C to demonstrate the robustness of the proposed approach in comparison with other state-of-the-art approaches.


We compare our approach with state-of-the-art joint image filtering approaches, including Joint Bilateral Filtering (JBF)\cite{kopf2007joint}, Guided image Filtering (GF)\cite{he2010guided}, Static/Dynamic Filtering (SDF)\cite{ham2017robust}, Deep Joint image Filtering (DJF)\cite{li2016deep} and Joint Filtering via optimizing a Scale Map (JFSM)\cite{shen2015multispectral} where the same RGB guidance images as in our approach are leveraged.
Our approach is also compared with several representative single image SR approaches, such as A+~\cite{timofte2014a+}, ANR (Anchored neighbourhood regression)~\cite{timofte2013anchored}, and the sparse coding algorithm of Zeyde \textit{et al.}\cite{zeyde2010single}.
%
Furthermore, we select bicubic interpolation as the baseline method. We adopt the Peak Signal to Noise Ratio (PSNR), the root-mean-square error (RMSE)
and the Structure SIMilarity (SSIM) index\cite{wang2004image}
as the image quality evaluation metrics which are commonly used in the image processing literature. 
The multi-spectral/RGB datasets are obtained from the Columbia multi-spectral database\footnote{\url{http://www.cs.columbia.edu/CAVE/databases/multispectral/}}. The infrared/RGB images datasets are obtained from the EPFL RGB-NIR Scene database\footnote{\url{http://ivrl.epfl.ch/supplementary_material/cvpr11/}}. All these datasets are registered for both modalities. 
For each multimodal dataset, we randomly separate its image pairs into two groups: training group and testing group. Then, we blur and downsample each HR image of target modality by a factor, e.g., 4 $\times$ and 6 $\times$, using the MATLAB "imresize" function to generate corresponding LR versions, similar to \cite{yang2008image,yang2010image}.

%
%


%
\begin{figure}[t]
	\centering
	\includegraphics[width = 8cm]{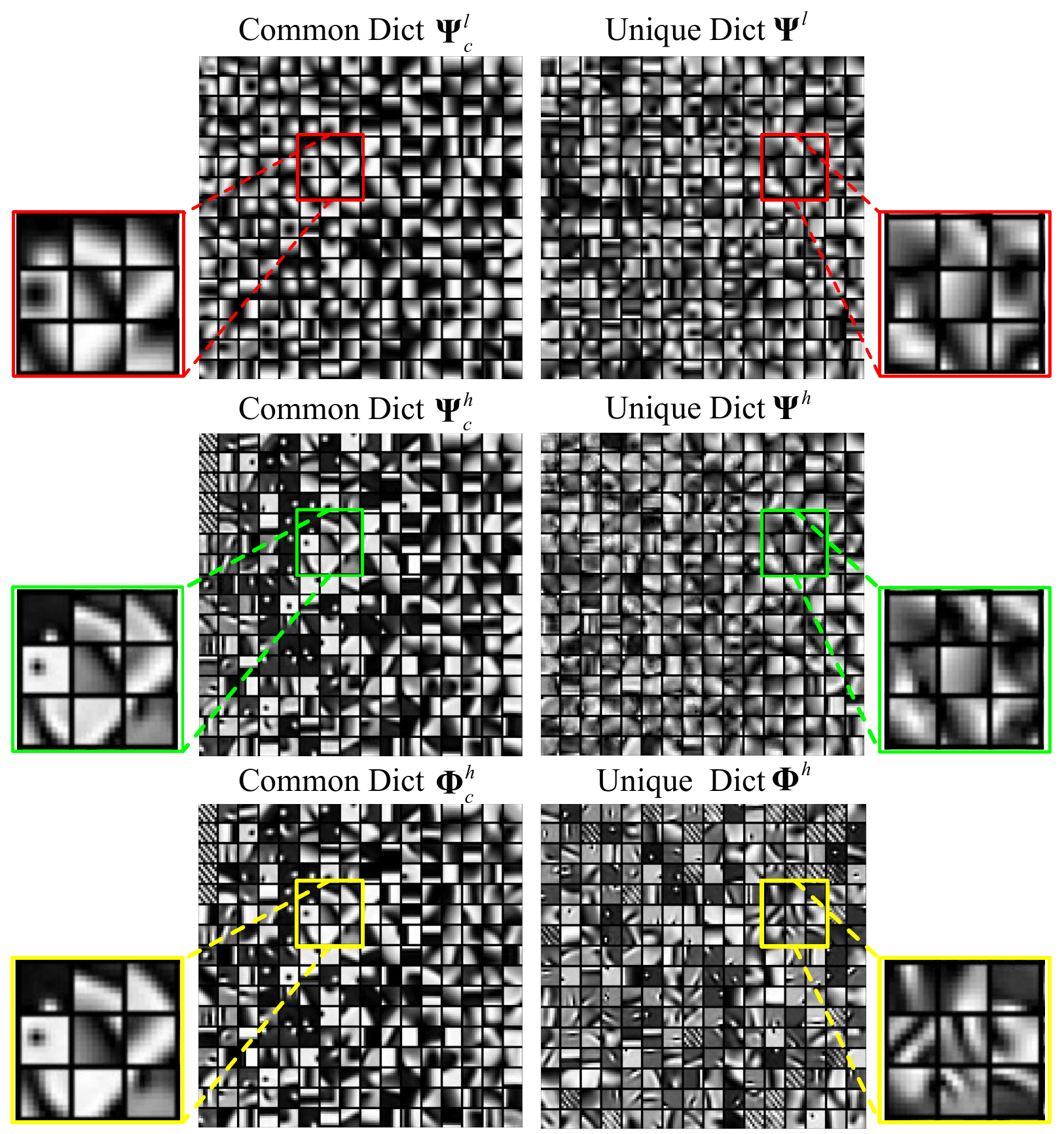} 
	%
	
	\caption{Learned coupled dictionaries for multi-spectral images of wavelength 640nm and RGB images. 256 atoms are shown here. The first row indicates the common and unique dictionaries learned for 4$\times$ downsampling LR multi-spectral images. The second row indicates the HR dictionary pair. The last row shows the dictionaries learned from side information, i.e. RGB images.}
	\label{Fig:CDL_PracticalResuts}
\end{figure}
%

\begin{figure*}[th]
	\centering
	\begin{minipage}[b]{0.11\linewidth}
		\centering
		{\footnotesize chart toy} \hfill 
		\includegraphics[width = 2cm, height= 2cm]{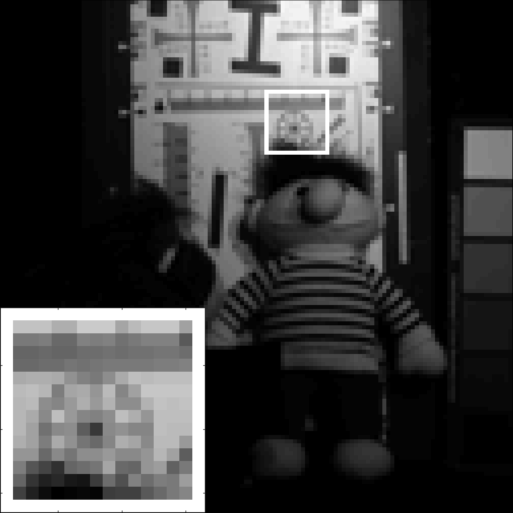}
	\end{minipage} 
	\begin{minipage}[b]{0.11\linewidth}
		\centering
		\includegraphics[width = 2cm, height= 2cm]{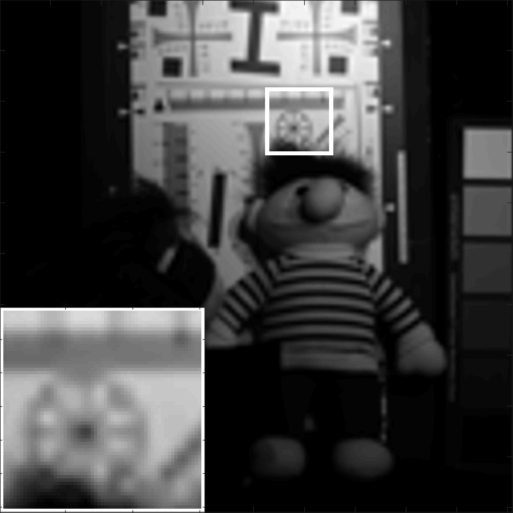}
	\end{minipage} 
	\begin{minipage}[b]{0.11\linewidth}
		\centering
		\includegraphics[width = 2cm, height= 2cm]{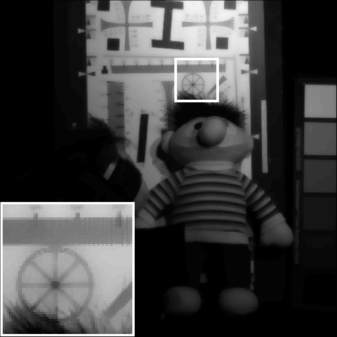}
	\end{minipage} 
	\begin{minipage}[b]{0.11\linewidth}
		\centering
		\includegraphics[width = 2cm, height= 2cm]{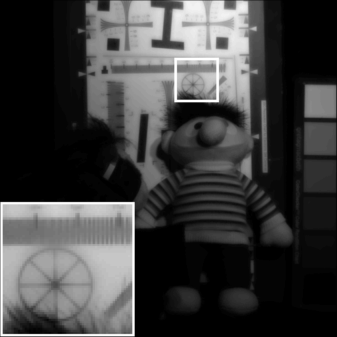}
	\end{minipage} 
	\begin{minipage}[b]{0.11\linewidth}
		\centering
		\includegraphics[width = 2cm, height= 2cm]{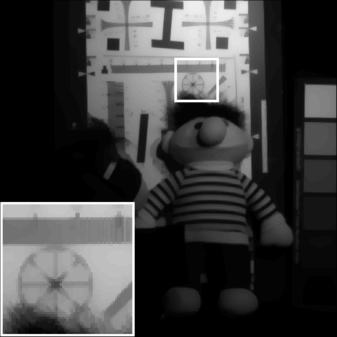}
	\end{minipage} 
	\begin{minipage}[b]{0.11\linewidth}
		\centering
		\includegraphics[width = 2cm, height= 2cm]{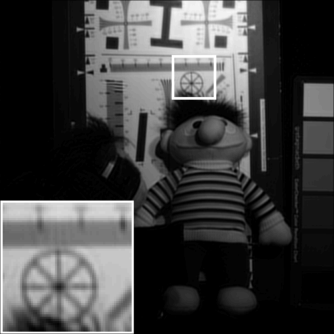}
	\end{minipage} 
	\begin{minipage}[b]{0.11\linewidth}
		\centering
		\includegraphics[width = 2cm, height= 2cm]{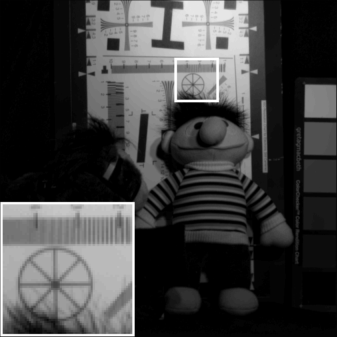}
	\end{minipage} 
	\begin{minipage}[b]{0.11\linewidth}
		\centering
		\includegraphics[width = 2cm, height= 2cm]{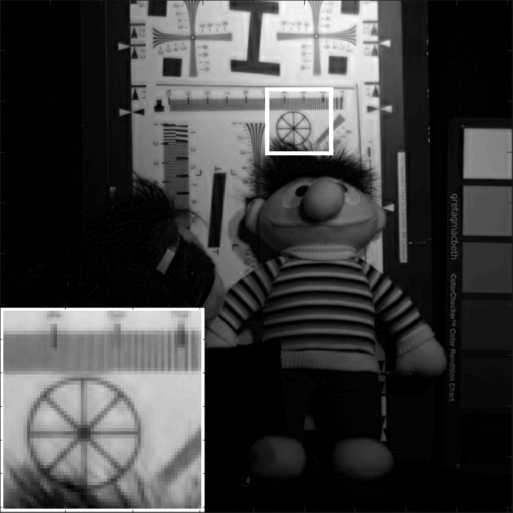}
	\end{minipage} 
	\\
	\begin{minipage}[b]{0.11\linewidth}
		\centering
		\includegraphics[width = 2cm, height= 2cm]{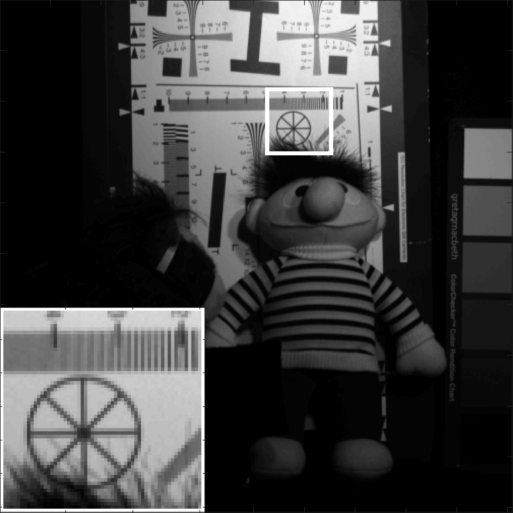}
	\end{minipage} 
	\begin{minipage}[b]{0.11\linewidth}
		\centering
		\includegraphics[width = 2cm, height= 2cm]{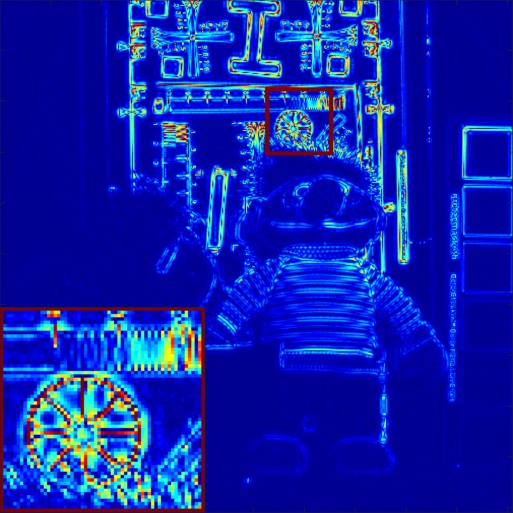}
	\end{minipage} 
	\begin{minipage}[b]{0.11\linewidth}
		\centering
		\includegraphics[width = 2cm, height= 2cm]{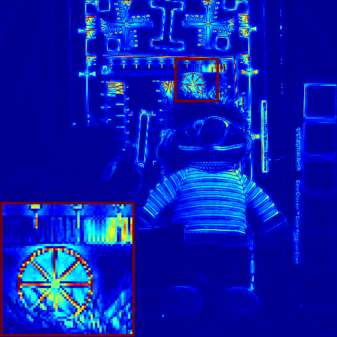}
	\end{minipage} 
	\begin{minipage}[b]{0.11\linewidth}
		\centering
		\includegraphics[width = 2cm, height= 2cm]{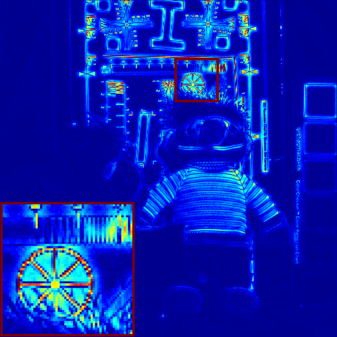}
	\end{minipage} 
	\begin{minipage}[b]{0.11\linewidth}
		\centering
		\includegraphics[width = 2cm, height= 2cm]{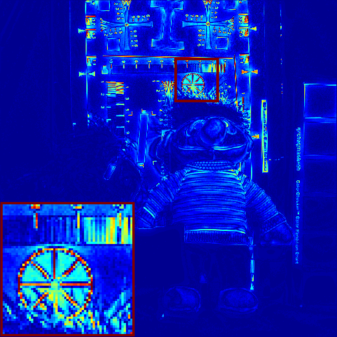}
	\end{minipage} 
	\begin{minipage}[b]{0.11\linewidth}
		\centering
		\includegraphics[width = 2cm, height= 2cm]{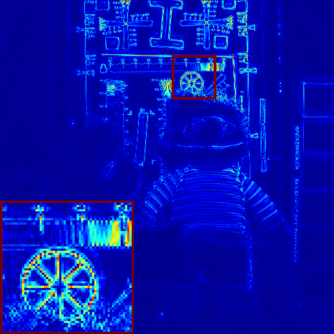}
	\end{minipage} 
	\begin{minipage}[b]{0.11\linewidth}
		\centering
		\includegraphics[width = 2cm, height= 2cm]{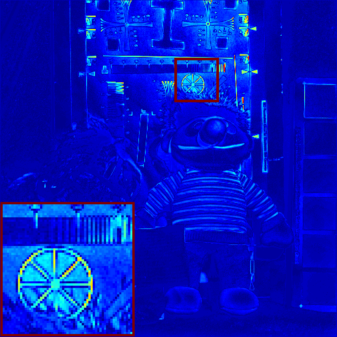}
	\end{minipage} 
	\begin{minipage}[b]{0.11\linewidth}
		\centering
		\includegraphics[width = 2cm, height= 2cm]{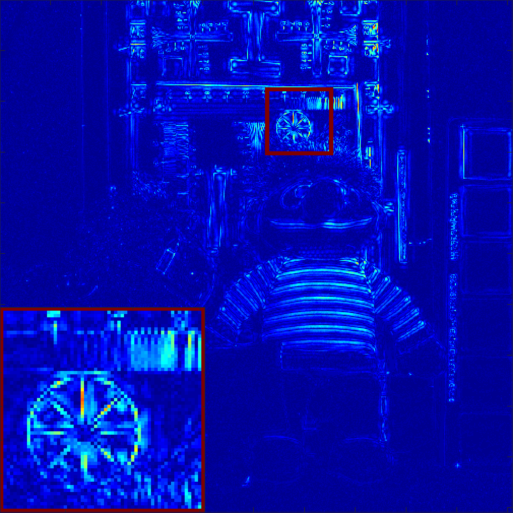}
	\end{minipage} 
	\\
	\begin{minipage}[b]{0.11\linewidth}
		\centering
		{\footnotesize feathers} \hfill 
		\includegraphics[width = 2cm, height= 2cm]{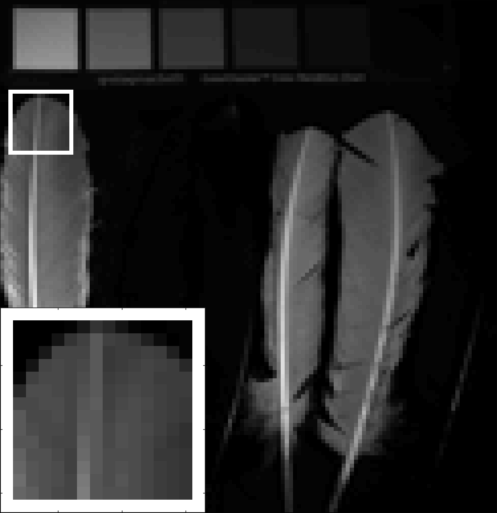}
	\end{minipage} 
	\begin{minipage}[b]{0.11\linewidth}
		\centering
		\includegraphics[width = 2cm, height= 2cm]{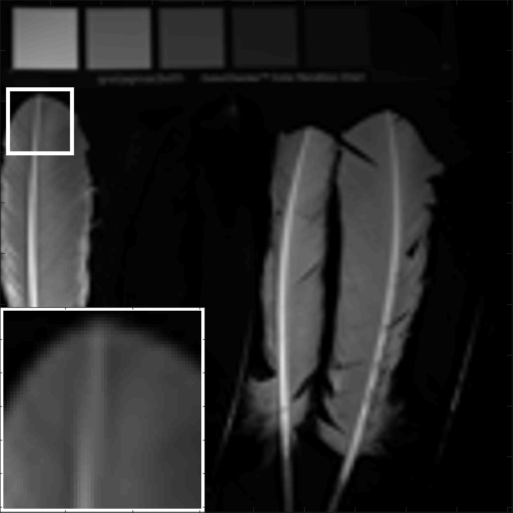}
	\end{minipage} 
	\begin{minipage}[b]{0.11\linewidth}
		\centering
		\includegraphics[width = 2cm, height= 2cm]{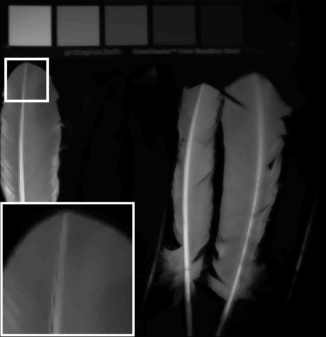}
	\end{minipage} 
	\begin{minipage}[b]{0.11\linewidth}
		\centering
		\includegraphics[width = 2cm, height= 2cm]{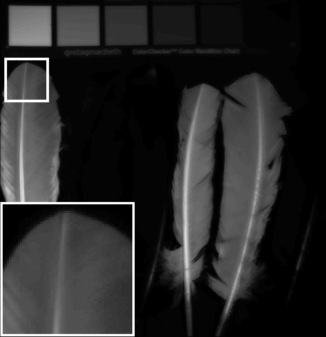}
	\end{minipage} 
	\begin{minipage}[b]{0.11\linewidth}
		\centering
		\includegraphics[width = 2cm, height= 2cm]{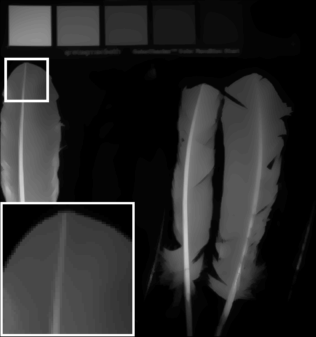}
	\end{minipage} 
	\begin{minipage}[b]{0.11\linewidth}
		\centering
		\includegraphics[width = 2cm, height= 2cm]{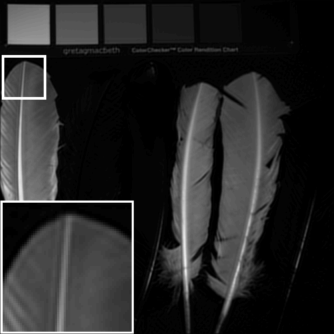}
	\end{minipage} 
	\begin{minipage}[b]{0.11\linewidth}
		\centering
		\includegraphics[width = 2cm, height= 2cm]{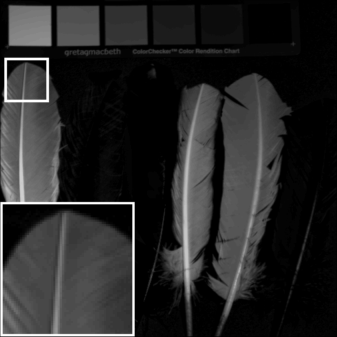}
	\end{minipage} 
	\begin{minipage}[b]{0.11\linewidth}
		\centering
		\includegraphics[width = 2cm, height= 2cm]{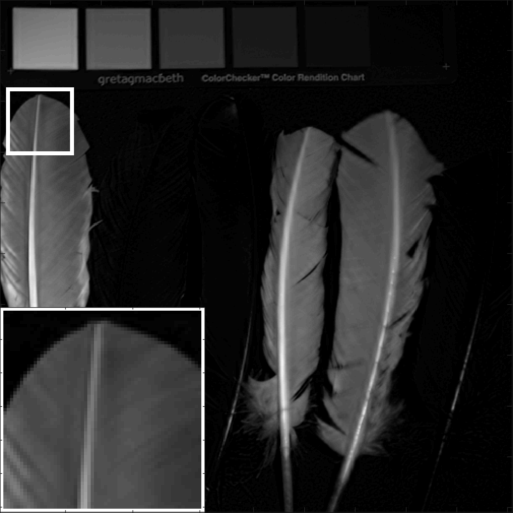}
	\end{minipage} 
	\\
	\begin{minipage}[b]{0.11\linewidth}
		\centering
		\includegraphics[width = 2cm, height= 2cm]{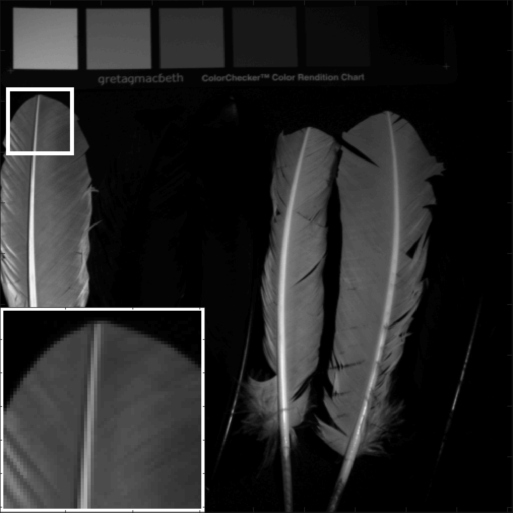}
	\end{minipage} 
	\begin{minipage}[b]{0.11\linewidth}
		\centering
		\includegraphics[width = 2cm, height= 2cm]{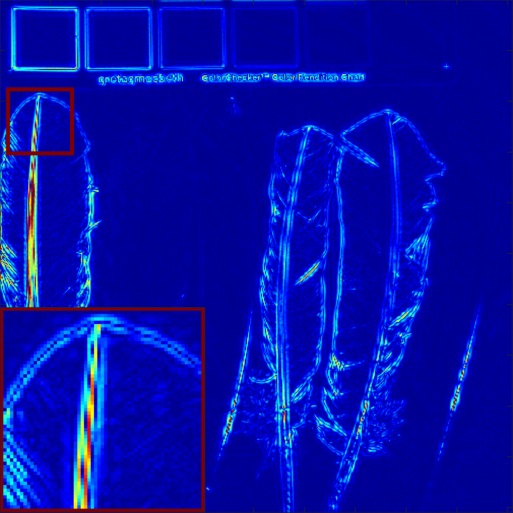}
	\end{minipage} 
	\begin{minipage}[b]{0.11\linewidth}
		\centering
		\includegraphics[width = 2cm, height= 2cm]{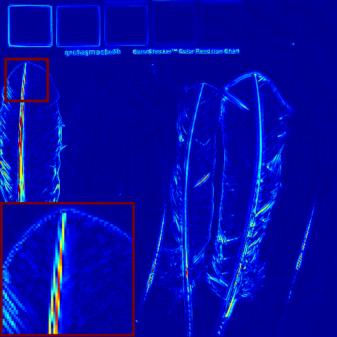}
	\end{minipage} 
	\begin{minipage}[b]{0.11\linewidth}
		\centering
		\includegraphics[width = 2cm, height= 2cm]{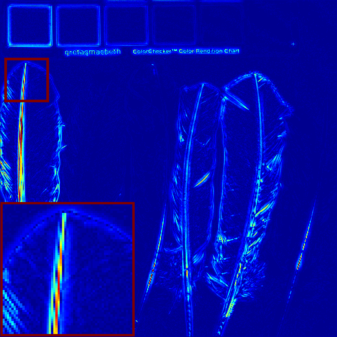}
	\end{minipage} 
	\begin{minipage}[b]{0.11\linewidth}
		\centering
		\includegraphics[width = 2cm, height= 2cm]{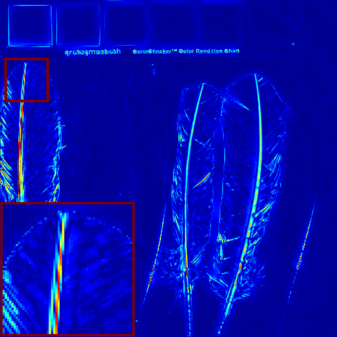}
	\end{minipage} 
	\begin{minipage}[b]{0.11\linewidth}
		\centering
		\includegraphics[width = 2cm, height= 2cm]{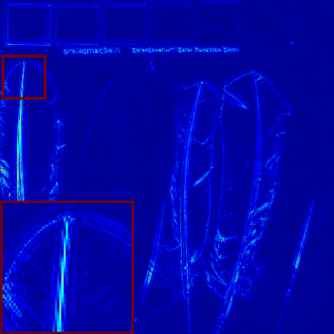}
	\end{minipage} 
	\begin{minipage}[b]{0.11\linewidth}
		\centering
		\includegraphics[width = 2cm, height= 2cm]{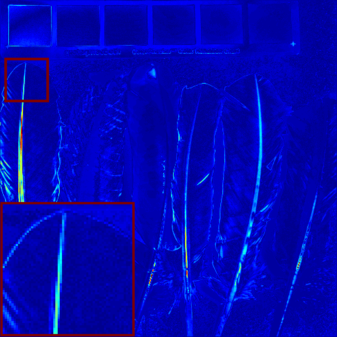}
	\end{minipage} 
	\begin{minipage}[b]{0.11\linewidth}
		\centering
		\includegraphics[width = 2cm, height= 2cm]{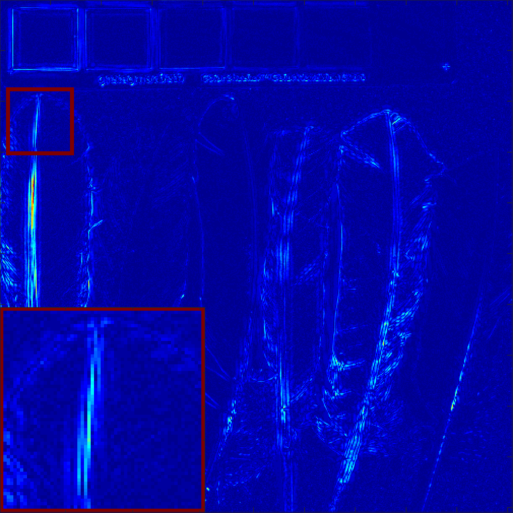}
	\end{minipage} 
	\\
	\begin{minipage}[b]{0.11\linewidth}
		\centering	{\footnotesize (a) Input/Truth} 
	\end{minipage}
	\begin{minipage}[b]{0.11\linewidth}
		\centering	{\footnotesize (b) Bicubic}
	\end{minipage} 
	\begin{minipage}[b]{0.11\linewidth}
		\centering	{\footnotesize (c) JBF\cite{kopf2007joint}} 
	\end{minipage} 
	\begin{minipage}[b]{0.11\linewidth}
		\centering	{\footnotesize (d) GF\cite{he2010guided}} 
	\end{minipage} 
	\begin{minipage}[b]{0.11\linewidth}
		\centering	{\footnotesize (e) SDF\cite{ham2017robust}} 
	\end{minipage} 
	\begin{minipage}[b]{0.11\linewidth}
		\centering	{\footnotesize (f) DJF\cite{li2016deep}} 
	\end{minipage}
	\begin{minipage}[b]{0.11\linewidth}
		\centering	{\footnotesize (g) JFSM\cite{shen2015multispectral}} 
	\end{minipage} 
	\begin{minipage}[b]{0.11\linewidth}
		\centering 	{\footnotesize (h) Proposed}
	\end{minipage} 
	\caption{4$\times$ upscaling for multi-spectral images of 640nm wavelength. For each image, the first row is the LR input and SR results. The second row is the ground truth and corresponding error map for each approach. In the error map, brighter area represents larger error.}
	\label{Fig:MS4x_Joint}
\end{figure*}

%
\begin{table*}[t]
	\centering
	\scriptsize
	\caption{4$\times$ upscaling for multi-spectral image of 640 nm band evaluated by PSNR (dB) and SSIM}
	\begin{tabular}{l| ll| ll| ll| ll| ll| ll |ll}
		\hline \hline
		& \multicolumn{2}{c|}{Bicubic} & \multicolumn{2}{c|}{JBF\cite{kopf2007joint}} & \multicolumn{2}{c|}{GF\cite{he2010guided}} & \multicolumn{2}{c|}{SDF\cite{ham2017robust}} & 
		\multicolumn{2}{c|}{DJF\cite{li2016deep}} & 
		\multicolumn{2}{c|}{JFSM\cite{shen2015multispectral}} & \multicolumn{2}{c}{Proposed} \\
		& SSIM & PSNR & SSIM & PSNR & SSIM & PSNR & SSIM & PSNR & SSIM & PSNR & SSIM & PSNR & SSIM & PSNR \\
		\hline
		chart toy & 0.9451 & 29.14 & 0.9528 & 30.69 & 0.9514 & 30.70 & 0.9523 & 30.74 & 0.9842 & 33.91 & 0.9215 & 33.300 & \textbf{0.9855} & \textbf{34.50} \\
		cloth & 0.7571 & 26.91 & 0.7640 & 27.62 & 0.7699 & 27.79 & 0.7315 & 27.18 & 0.9489 & 31.54 & \textbf{0.9770} & \textbf{35.330} & 0.9506 & 32.75 \\
		egyptian & 0.9761 & 36.22 & 0.9788 & 37.82 & 0.9788 & 37.96 & 0.9677 & 37.16 & 0.9861 & 41.31 & 0.9428 & 39.680 & \textbf{0.9935} & \textbf{42.63} \\
		feathers & 0.9530 & 30.46 & 0.9599 & 31.80 & 0.9618 & 32.12 & 0.9434 & 30.92 & 0.9848 & 36.01 & 0.9096 & 33.540 & \textbf{0.9871} & \textbf{36.25} \\
		glass tiles & 0.9215 & 26.38 & 0.9339 & 27.15 & 0.9326 & 27.45 & 0.9188 & 27.01 & \textbf{0.9814} & \textbf{31.83} & 0.9407 & 29.340 & 0.9791 & 31.05 \\
		jelly beans & 0.9269 & 27.45 & 0.9474 & 28.97 & 0.9488 & 29.54 & 0.9279 & 27.87 & 0.9820 & 32.77 & 0.9356 & 30.820 & \textbf{0.9866} & \textbf{34.38} \\
		oil painting & 0.9025 & 32.23 & 0.9034 & 33.23 & 0.9033 & 33.30 & 0.9001 & 32.80 & 0.9493 & 34.39 & 0.9439 & 34.160 & \textbf{0.9601} & \textbf{36.24} \\
		paints & 0.9569 & 30.47 & 0.9714 & 32.08 & 0.9698 & 32.23 & 0.9569 & 31.35 & 0.9897 & \textbf{37.74} & 0.9321 & 32.960 & \textbf{0.9900} & 36.99 \\
		\hline
		average & 0.9174 & 29.91 & 0.9265 & 31.17 & 0.9270 & 31.39 & 0.9123 & 30.63 & 0.9758 & 34.94 & 0.9379 & 33.640 & \textbf{0.9791} & \textbf{35.60} \\
		\hline \hline
	\end{tabular}
	\label{Tab:MS4x_Joint}
\end{table*}

\subsection{Multi-spectral image SR}
\label{ssec:MS-SR}


\mypar{Training Phase with CDL}
Before the coupled dictionary learning, we adopt some common preprocessing operations. Specifically, we upscale the LR multi-spectral training images to the desired size (i.e. the same size as HR version) using bicubic interpolation. The RGB images are converted from RGB to YCbCr space where we only use the luminance channel as the guidance, since human eyes are more sensitive to luminance information than chrominance information. Then, the interpolated LR images, the target HR images and the corresponding guidance images are divided into a set of $\sqrt{N} \times \sqrt{N}$ patch pairs.
We remove the mean from each patch, as the DC component is always preserved well during the upscaling process. Then, we vectorize the patches to form the training datasets $\mathbf{X}^l$, $\mathbf{X}^h$ and $\mathbf{Y}$ of dimension $N \times T$. Smooth patches with variance less than 0.02 have been eliminated as they are less informative. Once the training dataset is prepared, we apply our coupled dictionary learning algorithm, shown in Algorithm~\ref{Alg:CK-SVD}, to learn the dictionary pairs $[\boldsymbol{\Psi}_{c}^l,\boldsymbol{\Psi}^l]$ and  $[\boldsymbol{\Phi}_{c},\boldsymbol{\Phi}]$ from $\mathbf{X}^l$ and $\mathbf{Y}$. Then, HR dictionary pair $[\boldsymbol{\Psi}_{c}^h,\boldsymbol{\Psi}^h]$ are computed based on $\mathbf{X}^h$ and the acquired sparse codes $\mathbf{Z}$ and $\mathbf{U}$. The parameter setting is as follows: patch size $\sqrt{N} \times \sqrt{N} = 8 \times 8$ for 4$\times$ upscaling and $16 \times 16$ for 6$\times$ upscaling, dictionary size $K= 1024$, total sparsity constraint $s = 20$, training size $T \approx 15,000$.

%

\begin{table*}[th]
	\centering
	\scriptsize
	\caption{6$\times$ upscaling for multi-spectral image of 640 nm band evaluated by PSNR (dB) and SSIM}
	\begin{tabular}{l| ll| ll| ll| ll| ll| ll |ll}
		\hline \hline
		& \multicolumn{2}{c|}{Bicubic} & \multicolumn{2}{c|}{JBF\cite{kopf2007joint}} & \multicolumn{2}{c|}{GF\cite{he2010guided}} & \multicolumn{2}{c|}{SDF\cite{ham2017robust}} & 
		\multicolumn{2}{c|}{DJF\cite{li2016deep}} & 
		\multicolumn{2}{c|}{JFSM\cite{shen2015multispectral}} & \multicolumn{2}{c}{Proposed} \\
		& SSIM & PSNR & SSIM & PSNR & SSIM & PSNR & SSIM & PSNR & SSIM & PSNR & SSIM & PSNR & SSIM & PSNR \\
		\hline
		chart toy & 0.8774 & 26.83 & 0.8992 & 28.08 & 0.8932 & 27.86 & 0.9006 & 28.12 & \textbf{0.9772} & \textbf{32.61} & 0.9144 & 31.22 & 0.9682 & 32.55 \\
		cloth & 0.6143 & 25.55 & 0.6424 & 26.06 & 0.6394 & 26.07 & 0.6158 & 25.80 & 0.9226 & 30.09 & \textbf{0.9723} & \textbf{33.79} & 0.9256 & 31.73 \\
		egyptian & 0.9459 & 33.79 & 0.9560 & 34.95 & 0.9536 & 34.80 & 0.9466 & 34.83 & 0.9681 & 40.24 & 0.9444 & 38.43 & \textbf{0.9872} & \textbf{40.75} \\
		feathers & 0.8973 & 27.68 & 0.9177 & 28.80 & 0.9138 & 28.76 & 0.9062 & 28.50 & \textbf{0.9765} & \textbf{34.09} & 0.9042 & 31.32 & 0.9727 & 33.75 \\
		glass tiles & 0.8401 & 24.45 & 0.8652 & 25.05 & 0.8585 & 25.06 & 0.8556 & 25.03 & \textbf{0.9705} & \textbf{30.33} & 0.9233 & 27.33 & 0.9646 & 29.87 \\
		jelly beans & 0.8424 & 24.93 & 0.8835 & 26.24 & 0.8801 & 26.36 & 0.8681 & 25.60 & 0.9721 & 31.22 & 0.9225 & 28.58 & \textbf{0.9734} & \textbf{32.73} \\
		oil painting & 0.8511 & 30.90 & 0.8664 & 31.87 & 0.8626 & 31.78 & 0.8574 & 31.49 & 0.9392 & 33.77 & \textbf{0.9462} & 34.09 & 0.9427 & \textbf{35.18} \\
		paints & 0.9005 & 27.51 & 0.9328 & 29.04 & 0.9253 & 28.90 & 0.9226 & 28.69 & \textbf{0.9842} & \textbf{36.60} & 0.9363 & 31.25 & 0.9792 & 34.93 \\
		\hline
		average & 0.8461 & 27.70 & 0.8704 & 28.76 & 0.8658 & 28.70 & 0.8591 & 28.50 & 0.9638 & 33.62 & 0.9330 & 32.00 & \textbf{0.9642} & \textbf{33.93} \\
		\hline \hline
	\end{tabular}
	\label{Tab:MS6x_Joint}
\end{table*}

Figure~\ref{Fig:CDL_PracticalResuts} shows the learned coupled dictionaries for multi-spectral images of wavelength 640 nm and the corresponding RGB version.  We can find that any pair of LR and HR atoms from $\boldsymbol{\Psi}_{c}^l$ and $\boldsymbol{\Psi}_{c}^h$ capture associated edges, blobs, textures with the same direction and location. Similar behavior can also be observed in $\boldsymbol{\Psi}^l$ and $\boldsymbol{\Psi}^h$. This implies that LR and HR dictionaries are indeed closely related to each other. On the other hand, LR and HR atom pairs also exhibit some differences. Specifically, the edges and textures captured by LR atoms tend to be blurred and smoothed, while they tend to be clearer and sharper in the corresponding HR atoms. More importantly, the common dictionary $\boldsymbol{\Phi}_{c}^h$ from the guidance images exhibits considerable resemblance and strong correlation to $\boldsymbol{\Psi}_{c}^h$ and $\boldsymbol{\Psi}_{c}^l$ from the HR/LR modalities of interest. This indicates that the three common dictionaries have indeed captured the similarities between multi-spectral and RGB modalities. In contrast, the learned unique dictionaries $\boldsymbol{\Psi}^h$ and $\boldsymbol{\Phi}$ represent the disparities of these modalities and therefore rarely exhibit resemblance.

\mypar{Testing Phase with CSR}
During the coupled super-resolution phase, given a new pair of LR multi-spectral and HR RGB images for test, we upscale the LR multi-spectral image to the desired size as before. Then the testing image pair are subdivided into overlapping patches of size $\sqrt{N} \times \sqrt{N}$ pixels with overlap stride equal to 1 pixel.\footnote{The overlap stride denotes the distance between corresponding pixel locations in adjacent image patches.} The DC component is also removed from each patch and stored. We vectorize these patches to construct the testing datasets $\mathbf{x}^l_{test}$ and $\mathbf{y}_{test}$. Then, we perform coupled sparse coding on $\mathbf{x}^l_{test}$ and $\mathbf{y}_{test}$ with respect to learned dictionary pairs $[\boldsymbol{\Psi}_{c}^l,\boldsymbol{\Psi}^l]$ and  $[\boldsymbol{\Phi}_{c},\boldsymbol{\Phi}]$ to obtain the approximated sparse codes $\mathbf{z}_{test}$, $\mathbf{u}_{test}$ and $\mathbf{v}_{test}$, which are then multiplied with the HR dictionary pair $[\boldsymbol{\Psi}_{c}^h,\boldsymbol{\Psi}^h]$ to predict the HR patches $\mathbf{x}^h_{test}$, shown in Algorithm~\ref{Alg:CSR}. Finally, the DC component of each patch is added back to the corresponding estimated HR patch. These HR patches are tiled together and the overlapping areas are averaged to reconstruct the HR image of interest.

Figure~\ref{Fig:MS4x_Joint} shows the multi-spectral image SR results for the 640 nm wavelength band. 
As we can see, the reconstructed MS image and its corresponding residual from bicubic interpolation, JBF\cite{kopf2007joint}, GF\cite{he2010guided}, SDF\cite{ham2017robust} and DJF\cite{li2016deep} exhibit noticeable blurred areas. The reconstruction from JFSM\cite{shen2015multispectral} shows sharp edges but with weaker intensity than the ground-truth, a form of luminance distortion resulting from texture copying artifacts (see the zoom-in area of the wheel in the chart toy). In comparison, our approach is able to reliably recover more accurate image details and, at the same time, substantially suppresses ringing artifacts. Therefore, our reconstruction is more photo-realistic and visually appealing than the counterparts. This is also confirmed by the error maps, as well as by quantitative measure in terms of PSNR and SSIM, shown in Table~\ref{Tab:MS4x_Joint} and Table~\ref{Tab:MS6x_Joint} for 4$\times$ and 6$\times$ upscaling, respectively.\footnote{
	Limited to space, only a few algorithms producing the best results are shown in the paper. More detailed results can be found in the supplementary materials.} 
The quantitative results show that our approach outperforms bicubic interpolation with significant gains of average 5.6dB, 6.2dB and also exhibits notable advantage over the state-of-the-art joint image filtering approaches. For both 4$\times$ and 6$\times$ upscaling, the proposed approach outperforms JBF\cite{kopf2007joint}, GF\cite{he2010guided}, SDF\cite{ham2017robust}, JFSM\cite{shen2015multispectral} with gains of at least 1.9dB in terms of average PSNR. Our approach also outperforms the deep-learning-based approach DJF\cite{li2016deep} for the selected number of training samples.

\vspace{-0.3cm}

\subsection{Near-infrared image SR}
\label{ssec:NIR-SR}
We also evaluate our approach on near-infrared (NIR) images with registered RGB images as side information. As the response of NIR band has poor correlation with the response of the visible band, it is usually difficult to infer the brightness of a NIR image given a corresponding RGB modality. Thus, it is more challenging to take good advantage of the RGB version to super-resolve the near-infrared version. The LR/HR training/testing dataset and the side information are prepared in a manner similar to the previous multi-spectral case. The parameter setting keeps the same as before.
The first dataset includes houses and buildings that contain many fine textures and sharp edges. This makes the SR task more challenging than super-resolving images with smoother textures. The second dataset includes natural landscape images with water, trees, stone and more.

\begin{figure*}[t]
	\centering
	\begin{minipage}[b]{0.11\linewidth}
		\centering
		{\footnotesize urban\_0004} \hfill 
		\includegraphics[width = 2cm, height= 2.1cm]{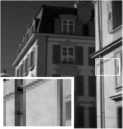}
	\end{minipage} 
	\begin{minipage}[b]{0.11\linewidth}
		\centering
		\includegraphics[width = 2cm, height= 2.1cm]{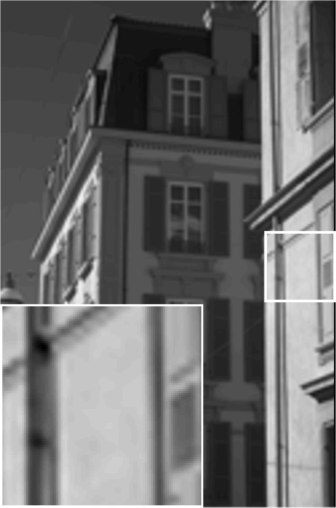}
	\end{minipage} 
	\begin{minipage}[b]{0.11\linewidth}
		\centering
		\includegraphics[width = 2cm, height= 2.1cm]{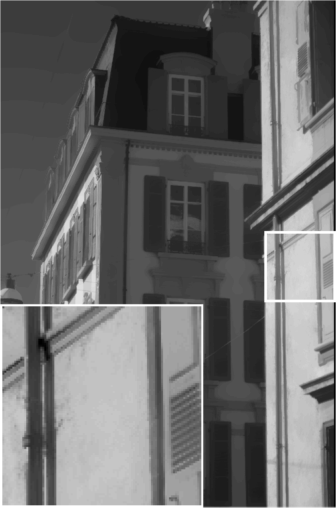}
	\end{minipage} 
	\begin{minipage}[b]{0.11\linewidth}
		\centering
		\includegraphics[width = 2cm, height= 2.1cm]{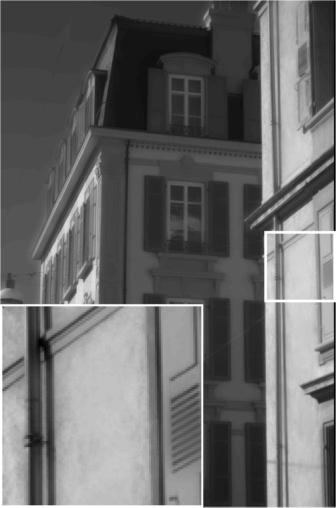}
	\end{minipage} 
	\begin{minipage}[b]{0.11\linewidth}
		\centering
		\includegraphics[width = 2cm, height= 2.1cm]{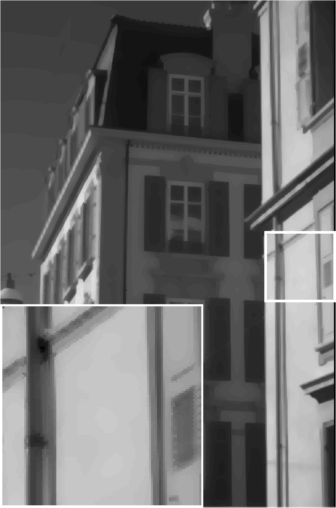}
	\end{minipage} 
	\begin{minipage}[b]{0.11\linewidth}
		\centering
		\includegraphics[width = 2cm, height= 2.1cm]{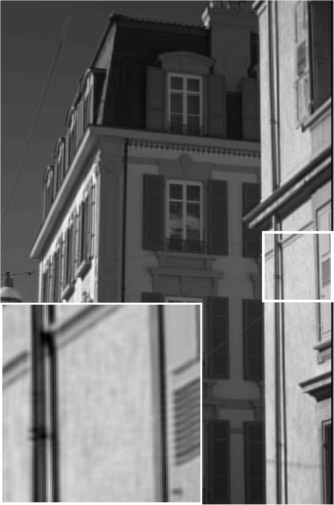}
	\end{minipage} 
	\begin{minipage}[b]{0.11\linewidth}
		\centering
		\includegraphics[width = 2cm, height= 2.1cm]{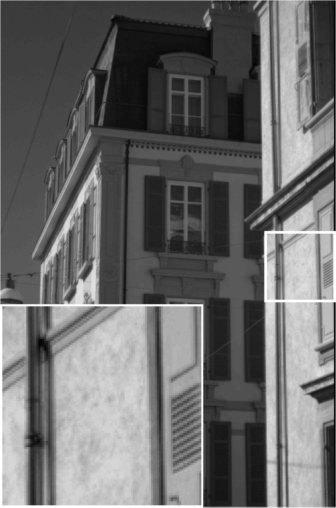}
	\end{minipage} 
	\begin{minipage}[b]{0.11\linewidth}
		\centering
		\includegraphics[width = 2cm, height= 2.1cm]{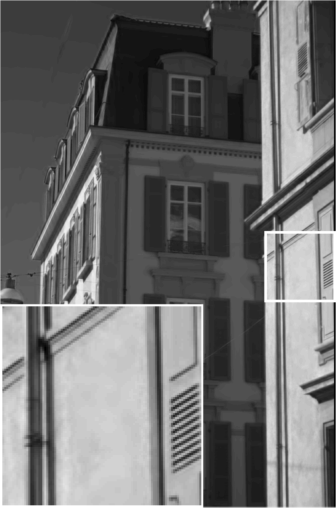}
	\end{minipage} 
	\\
	\begin{minipage}[b]{0.11\linewidth}
		\centering
		\includegraphics[width = 2cm, height= 2.1cm]{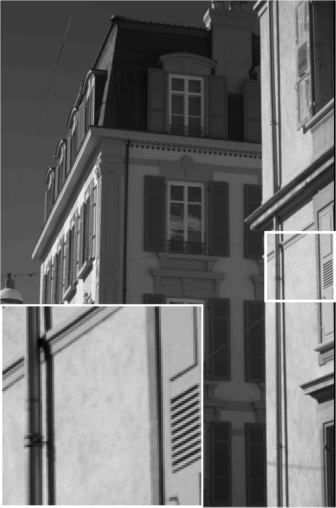}
	\end{minipage} 
	\begin{minipage}[b]{0.11\linewidth}
		\centering
		\includegraphics[width = 2cm, height= 2.1cm]{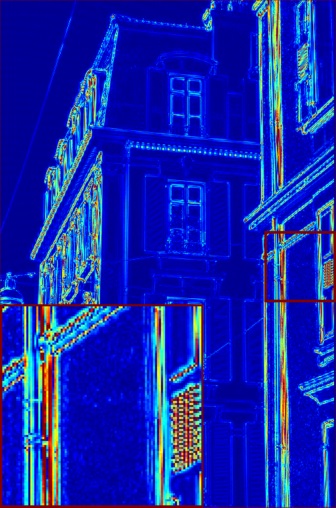}
	\end{minipage} 
	\begin{minipage}[b]{0.11\linewidth}
		\centering
		\includegraphics[width = 2cm, height= 2.1cm]{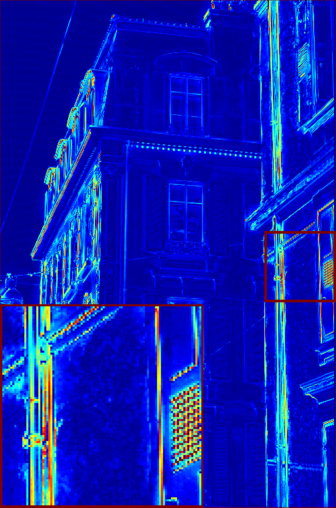}
	\end{minipage} 
	\begin{minipage}[b]{0.11\linewidth}
		\centering
		\includegraphics[width = 2cm, height= 2.1cm]{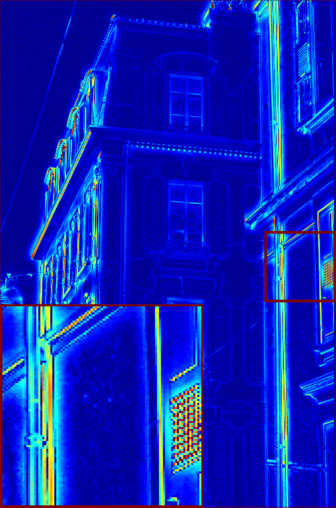}
	\end{minipage} 
	\begin{minipage}[b]{0.11\linewidth}
		\centering
		\includegraphics[width = 2cm, height= 2.1cm]{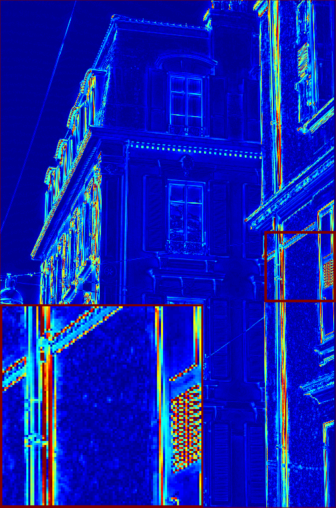}
	\end{minipage} 
	\begin{minipage}[b]{0.11\linewidth}
		\centering
		\includegraphics[width = 2cm, height= 2.1cm]{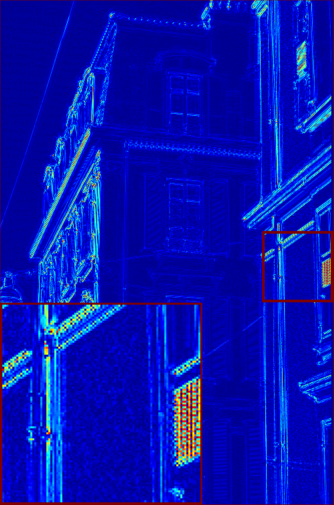}
	\end{minipage} 
	\begin{minipage}[b]{0.11\linewidth}
		\centering
		\includegraphics[width = 2cm, height= 2.1cm]{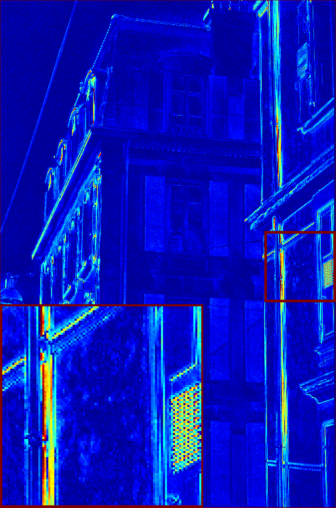}
	\end{minipage} 
	\begin{minipage}[b]{0.11\linewidth}
		\centering
		\includegraphics[width = 2cm, height= 2.1cm]{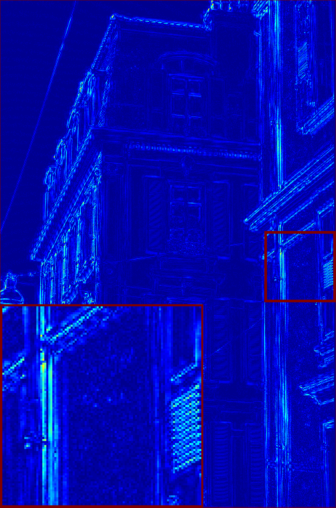}
	\end{minipage} 
	\\
	\begin{minipage}[b]{0.11\linewidth}
		\centering
		{\footnotesize urban\_0030} \hfill 
		\includegraphics[width = 2cm, height= 2.1cm]{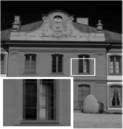}
	\end{minipage} 
	\begin{minipage}[b]{0.11\linewidth}
		\centering
		\includegraphics[width = 2cm, height= 2.1cm]{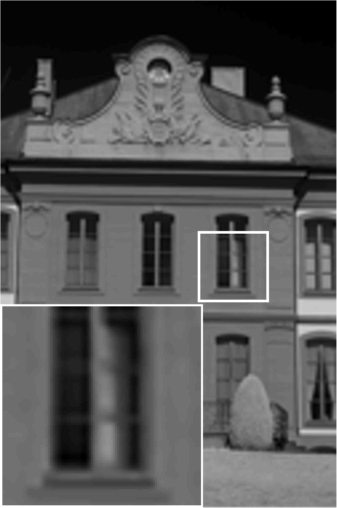}
	\end{minipage} 
	\begin{minipage}[b]{0.11\linewidth}
		\centering
		\includegraphics[width = 2cm, height= 2.1cm]{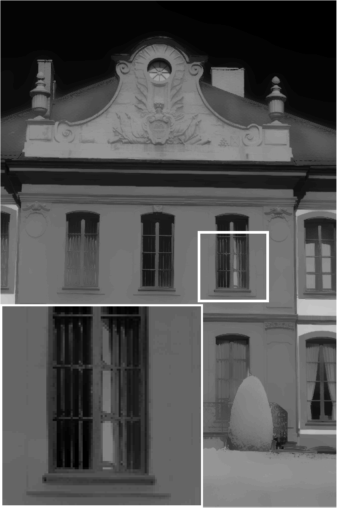}
	\end{minipage} 
	\begin{minipage}[b]{0.11\linewidth}
		\centering
		\includegraphics[width = 2cm, height= 2.1cm]{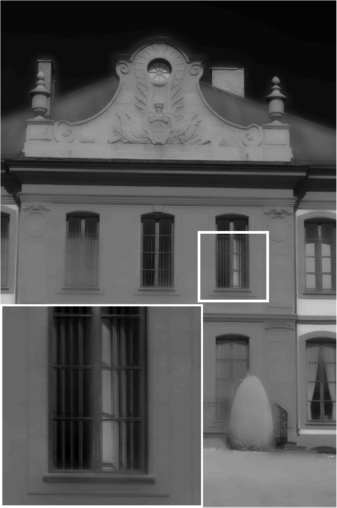}
	\end{minipage} 
	\begin{minipage}[b]{0.11\linewidth}
		\centering
		\includegraphics[width = 2cm, height= 2.1cm]{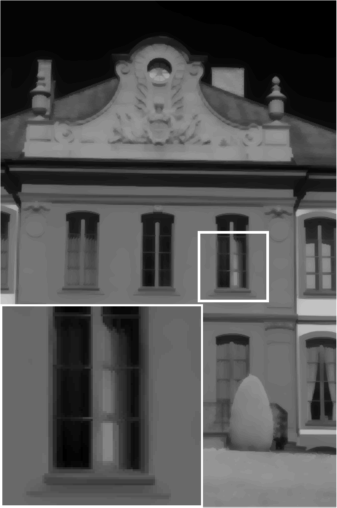}
	\end{minipage} 
	\begin{minipage}[b]{0.11\linewidth}
		\centering
		\includegraphics[width = 2cm, height= 2.1cm]{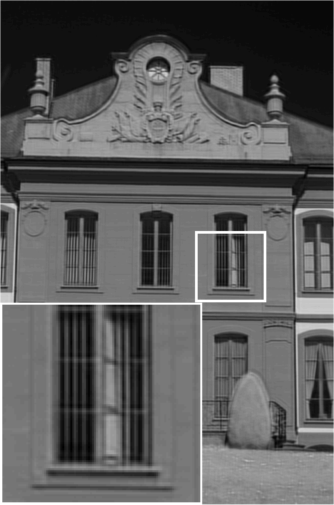}
	\end{minipage} 
	\begin{minipage}[b]{0.11\linewidth}
		\centering
		\includegraphics[width = 2cm, height= 2.1cm]{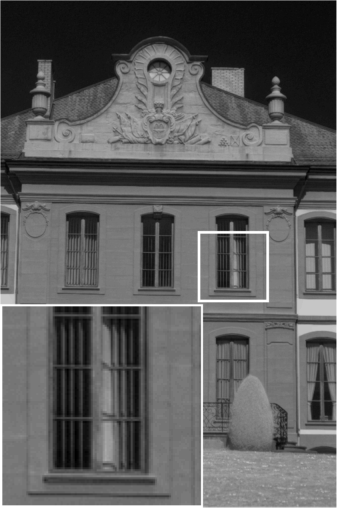}
	\end{minipage} 
	\begin{minipage}[b]{0.11\linewidth}
		\centering
		\includegraphics[width = 2cm, height= 2.1cm]{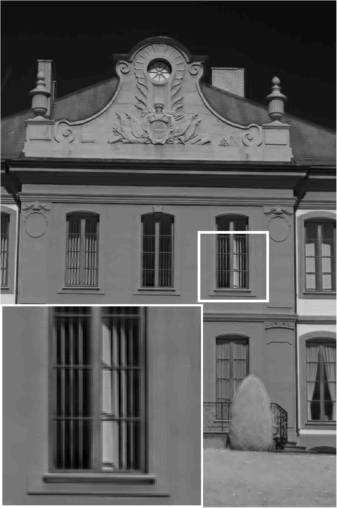}
	\end{minipage} 
	\\
	\begin{minipage}[b]{0.11\linewidth}
		\centering
		\includegraphics[width = 2cm, height= 2.1cm]{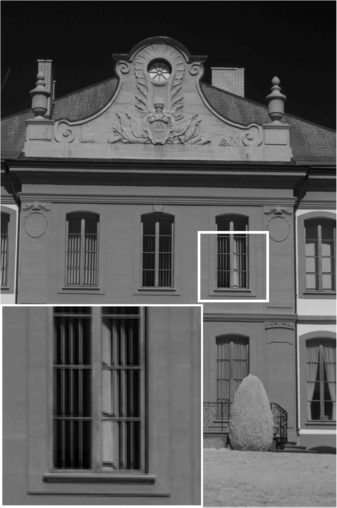}
	\end{minipage} 
	\begin{minipage}[b]{0.11\linewidth}
		\centering
		\includegraphics[width = 2cm, height= 2.1cm]{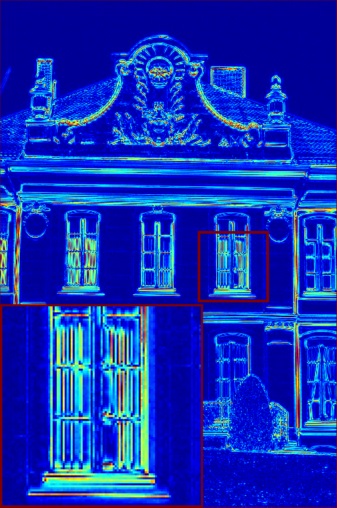}
	\end{minipage} 
	\begin{minipage}[b]{0.11\linewidth}
		\centering
		\includegraphics[width = 2cm, height= 2.1cm]{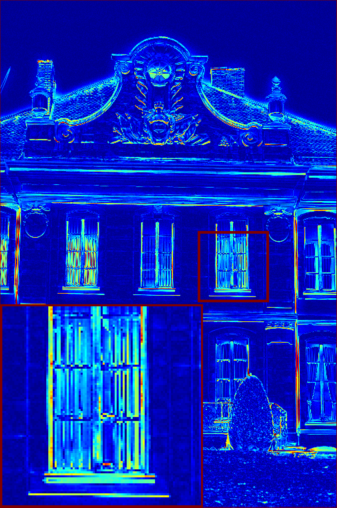}
	\end{minipage} 
	\begin{minipage}[b]{0.11\linewidth}
		\centering
		\includegraphics[width = 2cm, height= 2.1cm]{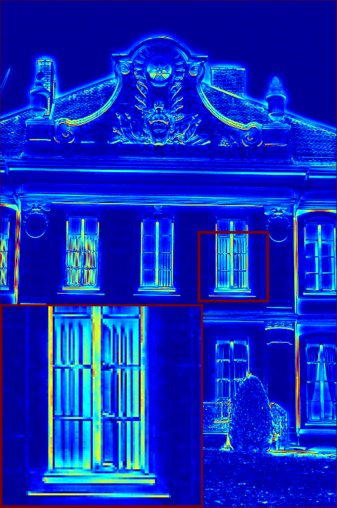}
	\end{minipage} 
	\begin{minipage}[b]{0.11\linewidth}
		\centering
		\includegraphics[width = 2cm, height= 2.1cm]{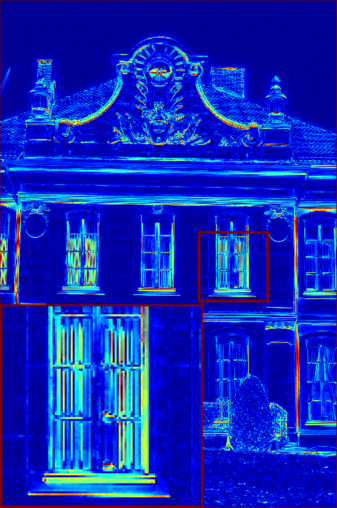}
	\end{minipage} 
	\begin{minipage}[b]{0.11\linewidth}
		\centering
		\includegraphics[width = 2cm, height= 2.1cm]{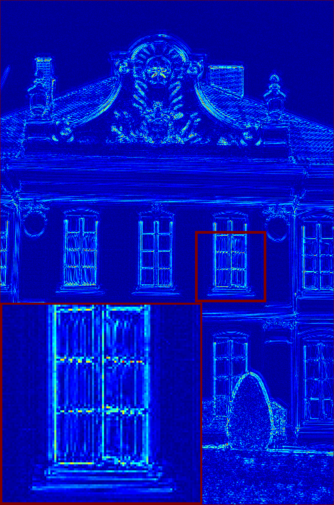}
	\end{minipage} 
	\begin{minipage}[b]{0.11\linewidth}
		\centering
		\includegraphics[width = 2cm, height= 2.1cm]{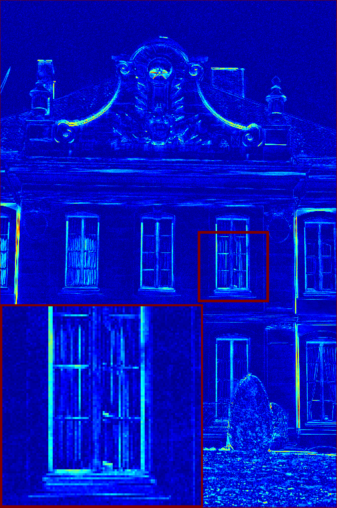}
	\end{minipage} 
	\begin{minipage}[b]{0.11\linewidth}
		\centering
		\includegraphics[width = 2cm, height= 2.1cm]{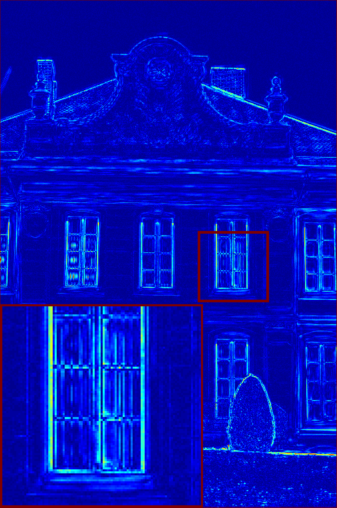}
	\end{minipage} 
	\\
	\begin{minipage}[b]{0.11\linewidth}
		\centering	{\footnotesize (a) Input/Truth} 
	\end{minipage}
	\begin{minipage}[b]{0.11\linewidth}
		\centering	{\footnotesize (b) Bicubic}
	\end{minipage} 
	\begin{minipage}[b]{0.11\linewidth}
		\centering	{\footnotesize (c) JBF\cite{kopf2007joint}} 
	\end{minipage} 
	\begin{minipage}[b]{0.11\linewidth}
		\centering	{\footnotesize (d) GF\cite{he2010guided}} 
	\end{minipage} 
	\begin{minipage}[b]{0.11\linewidth}
		\centering	{\footnotesize (e) SDF\cite{ham2017robust}} 
	\end{minipage} 
	\begin{minipage}[b]{0.11\linewidth}
		\centering	{\footnotesize (f) DJF\cite{li2016deep}} 
	\end{minipage} 
	\begin{minipage}[b]{0.11\linewidth}
		\centering	{\footnotesize (g) JFSM\cite{shen2015multispectral}} 
	\end{minipage} 
	\begin{minipage}[b]{0.11\linewidth}
		\centering 	{\footnotesize (h) Proposed}
	\end{minipage} 
	\caption{4$\times$ upscaling for near-infrared house images, e.g., urban\_0004(up) and urban\_0030(bottom). For each image, the first row is the LR input and SR results. The second row is the ground truth and corresponding error map for each approach. In the error map, brighter area represents larger error.}
	\label{Fig:NIR4x_Joint}
\end{figure*}

Figure~\ref{Fig:NIR4x_Joint} compares the visual quality of the reconstructed HR near-infrared images and the corresponding error maps. It can be seen that, on average, our approach recovers more visually plausible images, exhibiting less error than the competing methods.
Table~\ref{Tab:NIR4x_Joint} and~\ref{Tab:NIR6x_Joint} also confirm the significant advantage of the proposed approach over other state-of-the-art methods. In particular, this indicates that detailed structure information can be effectively captured by coupled dictionary learning, especially on images such as buildings and houses that contain a lot of sharp edges, textures and stripes.

\begin{figure}[th]
	\centering
	\begin{minipage}[b]{0.22\linewidth}
		\centering
		\includegraphics[width = 2cm, height= 2.5cm]{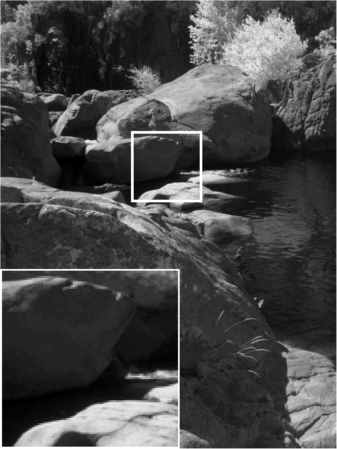}
	\end{minipage} 
	\begin{minipage}[b]{0.22\linewidth}
		\centering
		\includegraphics[width = 2cm, height= 2.5cm]{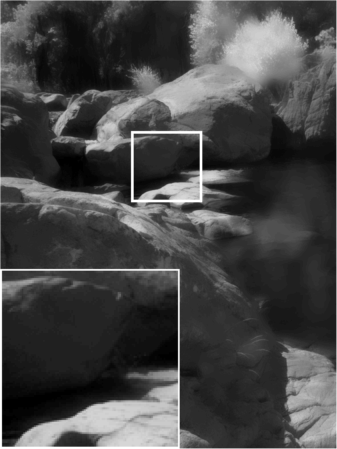}
	\end{minipage} 
	\begin{minipage}[b]{0.22\linewidth}
		\centering
		\includegraphics[width = 2cm, height= 2.5cm]{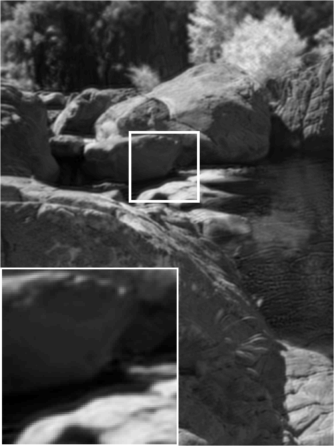}
	\end{minipage} 
	\begin{minipage}[b]{0.22\linewidth}
		\centering
		\includegraphics[width = 2cm, height= 2.5cm]{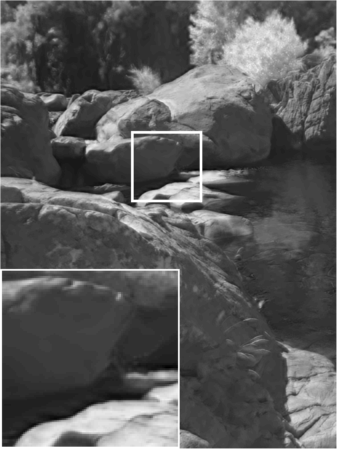}
	\end{minipage} 
	\\
	\begin{minipage}[b]{0.22\linewidth}
		\centering
		\includegraphics[width = 2cm, height= 2.5cm]{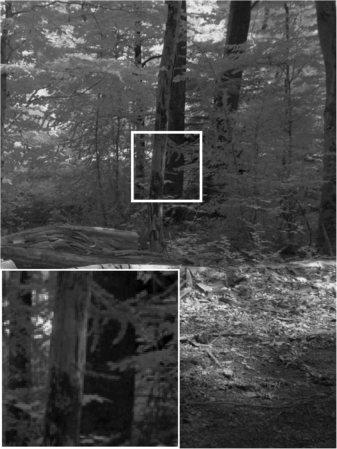}
	\end{minipage} 
	\begin{minipage}[b]{0.22\linewidth}
		\centering
		\includegraphics[width = 2cm, height= 2.5cm]{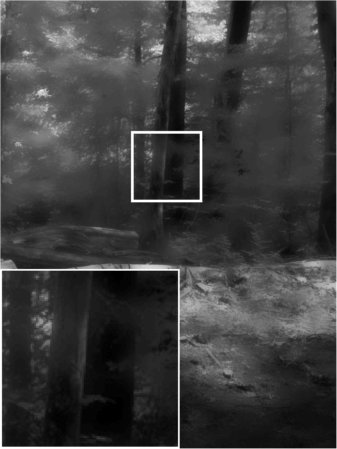}
	\end{minipage} 
	\begin{minipage}[b]{0.22\linewidth}
		\centering
		\includegraphics[width = 2cm, height= 2.5cm]{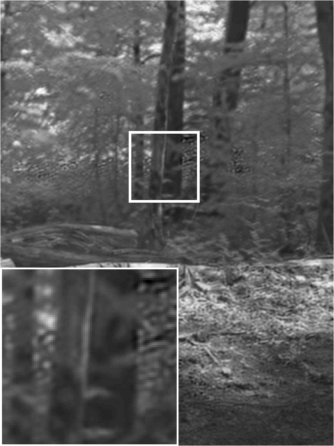}
	\end{minipage} 
	\begin{minipage}[b]{0.22\linewidth}
		\centering
		\includegraphics[width = 2cm, height= 2.5cm]{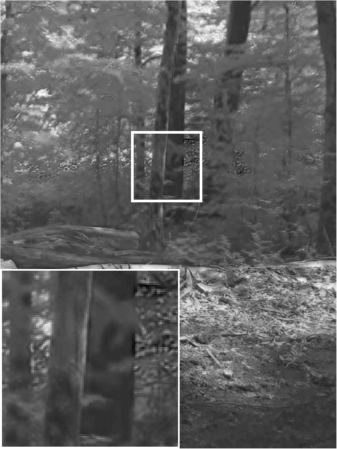}
	\end{minipage} 
	\\
	\begin{minipage}[b]{0.22\linewidth}
		\centering	{\footnotesize Truth} 
	\end{minipage}
	\begin{minipage}[b]{0.22\linewidth}
		\centering	{\footnotesize GF\cite{he2010guided}} 
	\end{minipage} 
	\begin{minipage}[b]{0.22\linewidth}
		\centering	{\footnotesize DJF\cite{li2016deep}} 
	\end{minipage} 
	\begin{minipage}[b]{0.22\linewidth}
		\centering 	{\footnotesize Proposed}
	\end{minipage} 
	\caption{4$\times$ upscaling for near-infrared landscape images, e.g., n0031(up) and n0051(bottom).}
	\label{Fig:NIR4x_Joint_Landscape}
\end{figure}

	Figure~\ref{Fig:NIR4x_Joint_Landscape} and Table~\ref{Tab:NIR4x_Joint_Landscape} show the visual and quantitative comparison for another dataset with landscape images.
	It can be seen that leaves, trees, grass and other natural objects with fine details tend to be over-smoothed in the reconstructed images from competing approaches. In contrast, these objects in our reconstruction appear clearer, sharper and less obscured. This further confirms the advantage of CDLSR in reliably restoring fine details without introducing notable artifacts.  (See additional comparisons with DJF\cite{li2016deep} in subsection \ref{sec:Experiments}-C.)

	Overall, the good performance of the proposed CDLSR approach is due to learned adaptive coupled dictionaries that are capable of effectively capturing salient features and critical correlations between the target and the guidance modalities in their sparse transform domains. These learned dictionaries can act as powerful priors that have the ability to dramatically reduce artifacts.

\begin{table*}[ht]
	\centering
	\scriptsize
	\caption{4$\times$ upscaling for near-infrared house images}
	\begin{tabular}{l| ll| ll| ll| ll| ll| ll |ll}
		\hline \hline
		& \multicolumn{2}{c|}{Bicubic} & \multicolumn{2}{c|}{JBF\cite{kopf2007joint}} & \multicolumn{2}{c|}{GF\cite{he2010guided}} & \multicolumn{2}{c|}{SDF\cite{ham2017robust}} & 
		\multicolumn{2}{c|}{DJF\cite{li2016deep}} & 
		\multicolumn{2}{c|}{JFSM\cite{shen2015multispectral}} & \multicolumn{2}{c}{Proposed} \\
		& SSIM & PSNR & SSIM & PSNR & SSIM & PSNR & SSIM & PSNR & SSIM & PSNR & SSIM & PSNR & SSIM & PSNR \\
		\hline
		urban\_0004 & 0.9029 & 25.93 & 0.9359 & 28.47 & 0.9391 & 28.75 & 0.9066 & 26.82 & 0.9789 & 31.02 & 0.9721 & 30.86 & \textbf{0.9811} & \textbf{34.14} \\
		urban\_0006 & 0.9458 & 30.89 & 0.9311 & 32.10 & 0.9400 & 32.66 & 0.8918 & 30.60 & \textbf{0.9894} & 36.04 & 0.9741 & 32.86 & 0.9868 & \textbf{36.79} \\
		urban\_0017 & 0.9527 & 30.45 & 0.9172 & 31.11 & 0.9205 & 31.32 & 0.9281 & 30.72 & \textbf{0.9815} & 34.18 & 0.9500 & 32.85 & 0.9777 & \textbf{35.27} \\
		urban\_0018 & 0.9298 & 25.19 & 0.9308 & 27.59 & 0.9251 & 27.70 & 0.9196 & 26.09 & \textbf{0.9888} & 30.72 & 0.9774 & 30.80 & 0.9874 & \textbf{33.01} \\
		urban\_0020 & 0.9577 & 28.03 & 0.9523 & 30.67 & 0.9494 & 30.69 & 0.9505 & 29.09 & \textbf{0.9915} & 33.60 & 0.9797 & 32.61 & 0.9893 & \textbf{36.66} \\
		urban\_0026 & 0.8704 & 26.27 & 0.8627 & 26.82 & 0.8571 & 26.89 & 0.8558 & 26.61 & 0.9397 & 29.21 & 0.9332 & 28.97 & \textbf{0.9482} & \textbf{30.35} \\
		urban\_0030 & 0.8401 & 26.54 & 0.8476 & 27.58 & 0.8383 & 27.59 & 0.8415 & 27.21 & 0.9345 & 31.27 & 0.9064 & 30.56 & \textbf{0.9443} & \textbf{32.71} \\
		urban\_0050 & 0.9434 & 26.65 & 0.9099 & 27.32 & 0.9116 & 27.35 & 0.9207 & 27.07 & 0.9616 & 28.58 & 0.9251 & 27.58 & \textbf{0.9663} & \textbf{29.37} \\
		\hline
		average & 0.9179 & 27.49 & 0.9109 & 28.96 & 0.9101 & 29.12 & 0.9018 & 28.03 & 0.9707 & 31.83 & 0.9522 & 30.89 & \textbf{0.9726} & \textbf{33.54} \\
		\hline \hline
	\end{tabular}
	\label{Tab:NIR4x_Joint}
\end{table*}

\begin{table*}[ht]
	\centering
	\scriptsize
	\caption{6$\times$ upscaling for near-infrared house images}
	\begin{tabular}{l| ll| ll| ll| ll| ll| ll |ll}
		\hline \hline
		& \multicolumn{2}{c|}{Bicubic} & \multicolumn{2}{c|}{JBF\cite{kopf2007joint}} & \multicolumn{2}{c|}{GF\cite{he2010guided}} & \multicolumn{2}{c|}{SDF\cite{ham2017robust}} & 
		\multicolumn{2}{c|}{DJF\cite{li2016deep}} & 
		\multicolumn{2}{c|}{JFSM\cite{shen2015multispectral}} & 	\multicolumn{2}{c}{Proposed} \\
		& SSIM & PSNR & SSIM & PSNR & SSIM & PSNR & SSIM & PSNR & SSIM & PSNR & SSIM & PSNR & SSIM & PSNR \\
		\hline
		urban\_0004 & 0.8094 & 23.87 & 0.8858 & 25.96 & 0.8817 & 25.94 & 0.8413 & 24.62 & \textbf{0.9670} & 29.68 & 0.9527 & 27.97 & 0.9558 & \textbf{30.77} \\
		urban\_0006 & 0.8671 & 28.48 & 0.8861 & 30.00 & 0.8876 & 30.17 & 0.8377 & 28.76 & \textbf{0.9830} & \textbf{34.92} & 0.9716 & 32.28 & 0.9664 & 34.15 \\
		urban\_0017 & 0.8998 & 28.64 & 0.8864 & 29.63 & 0.8860 & 29.61 & 0.8910 & 29.13 & \textbf{0.9599} & 32.80 & 0.9434 & 32.01 & 0.9515 & \textbf{32.98} \\
		urban\_0018 & 0.8393 & 23.07 & 0.8718 & 25.09 & 0.8591 & 24.98 & 0.8439 & 23.79 & \textbf{0.9844} & 29.92 & 0.9470 & 27.47 & 0.9727 & \textbf{31.03} \\
		urban\_0020 & 0.9053 & 26.03 & 0.9200 & 28.19 & 0.9118 & 28.01 & 0.9089 & 26.93 & \textbf{0.9873} & 32.61 & 0.9673 & 30.33 & 0.9763 & \textbf{33.85} \\
		urban\_0026 & 0.7850 & 24.71 & 0.8235 & 25.64 & 0.8131 & 25.63 & 0.7989 & 25.17 & \textbf{0.9183} & 28.38 & 0.9128 & 27.54 & 0.9172 & \textbf{28.88} \\
		urban\_0030 & 0.7517 & 25.19 & 0.7994 & 26.32 & 0.7855 & 26.22 & 0.7748 & 25.80 & 0.9063 & 30.00 & 0.8902 & 29.38 & \textbf{0.9099} & \textbf{30.52} \\
		urban\_0050 & 0.8921 & 25.17 & 0.8837 & 26.26 & 0.8846 & 26.26 & 0.8837 & 25.90 & \textbf{0.9414} & 27.64 & 0.9068 & 26.67 & 0.9402 & \textbf{28.37} \\
		\hline
		average & 0.8437 & 25.65 & 0.8696 & 27.13 & 0.8637 & 27.10 & 0.8475 & 26.26 & \textbf{0.9559} & 30.75 & 0.9365 & 29.21 & 0.9487 & \textbf{31.32} \\
		\hline \hline
	\end{tabular}
	\label{Tab:NIR6x_Joint}
\end{table*}

\begin{table}[t]
	\centering
	\scriptsize
	\caption{4$\times$ upscaling for near-infrared landscape images.}
	\label{Tab:NIR4x_Joint_Landscape}
	\begin{tabular}{C{0.6cm}| L{0.55cm}L{0.55cm}| L{0.55cm}L{0.55cm}| L{0.55cm}L{0.55cm}| L{0.55cm}L{0.55cm} }
		\hline \hline
		& \multicolumn{2}{c|}{GF\cite{he2010guided}} & \multicolumn{2}{c|}{JFSM\cite{shen2015multispectral}} & \multicolumn{2}{c|}{DJF\cite{li2016deep}} & \multicolumn{2}{c}{Proposed} \\
		& SSIM & PSNR & SSIM & PSNR & SSIM & PSNR & SSIM & PSNR \\
		\hline
		n0025 & 0.7970 & 27.14 & 0.8242 & 25.67 & 0.9093 & 28.43 & \textbf{0.9097} & \textbf{29.05} \\
		n0027 & 0.7002 & 25.82 & 0.7033 & 24.68 & 0.8565 & 27.87 & \textbf{0.8702} & \textbf{28.07} \\
		n0028 & 0.7519 & 25.01 & 0.7766 & 24.16 & \textbf{0.8812} & \textbf{26.88} & 0.8789 & 26.50 \\
		n0031 & 0.8524 & 27.81 & 0.8536 & 26.71 & {0.9111} & \textbf{28.72} & \textbf{0.9136} & 28.64 \\
		n0049 & 0.7832 & 29.52 & 0.7453 & 26.85 & \textbf{0.9021} & 31.49 & {0.8996} & \textbf{31.88} \\
		n0051 & 0.7262 & 25.97 & 0.7606 & 25.19 & 0.8732 & 27.64 & \textbf{0.8767} & \textbf{28.29} \\
		\hline
		average & 0.7685 & 26.88 & 0.7773 & 25.54 & 0.8889 & 28.50 & \textbf{0.8914} & \textbf{28.74}\\	
		\hline \hline 
	\end{tabular}
\end{table}

\vspace{-0.3cm}

\subsection{Proposed CDLSR vs Deep-Learning-Based SR Approaches}
\label{ssec:NIR-SR_Noise}
The previous experiments have shown that for a relatively modest number of training samples our proposed approach can lead to better results than the state-of-the-art, including deep-learning-based multimodal super-resolution methods (DJF~\cite{li2016deep}). 
However, as deep-learning-based methods can also successfully take advantage of the availability of a huge amount of data for training, DJF~\cite{li2016deep} eventually outperforms our approach, given enough training data. 

Our proposed approach has nonetheless other advantages with respect to DJF~\cite{li2016deep}. One advantage relates to the amount of training time required by DJF~\cite{li2016deep} in relation to CDLSR. For example, DJF\cite{li2016deep} takes about 12 hours to train through 50 epochs with an NVIDIA Titan black GPU for acceleration, while our approach takes only a few minutes for training a group of coupled dictionaries without any GPU acceleration. But our approach is slower than the deep-learning-based approach DJF during testing because we solve a non-convex optimization problem while DJF only performs a simple forward pass.

More importantly, the other advantage relates to the robustness of our approach in the presence of noise at training and/or testing stages, which is very common in practice~\cite{zhang2010two,nguyen2017bounded}. In particular, we repeat the previous NIR-SR experiments to test the robustness of both algorithms in the presence of contamination of additive zero-mean Gaussian noise. The training dataset for DJF~\cite{li2016deep} consists of 160,000 $33 \times 33$ patches and for the proposed CDLSR only 15000 patches of size $8\times8$ pixels. Each evaluation metric value is averaged on all the testing images. We consider two typical scenarios:\footnote{We assume that only the target modality is contaminated by noise and the guidance modality keeps clean as before in order to compare with previous noise-free situations.}


\subsubsection{LR noisy testing images}
\begin{figure}[t]
	\centering
	\begin{minipage}[b]{0.48\linewidth}
		\centering
		\includegraphics[width = 4.2cm]{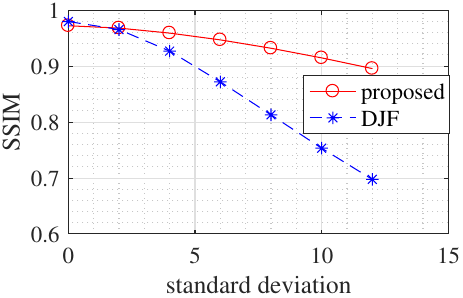} \\
		\scriptsize (a) SSIM w.r.t. Testing Noise
	\end{minipage} 
	\begin{minipage}[b]{0.48\linewidth}
		\centering
		\includegraphics[width = 4.2cm]{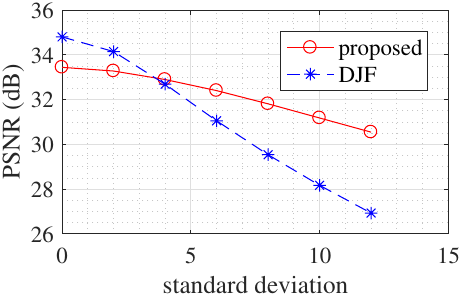} \\
		\scriptsize (b) PSNR w.r.t. Testing Noise
	\end{minipage}
	\caption{LR noisy testing images. The testing noise $\sigma_{test}$ ranges from 2 to 12 and the training noise $\sigma_{train} = 0$, for 4$\times$ upscaling of near-infrared house images using DJF~\cite{li2016deep} and proposed CDLSR.}
	\label{Fig:NIR4x_Joint_Noise_Line}
\end{figure}

\begin{figure}[t]
	\centering
	\begin{minipage}[b]{0.32\linewidth}
		\centering
		\includegraphics[width = 2.8cm, height= 2.1cm]{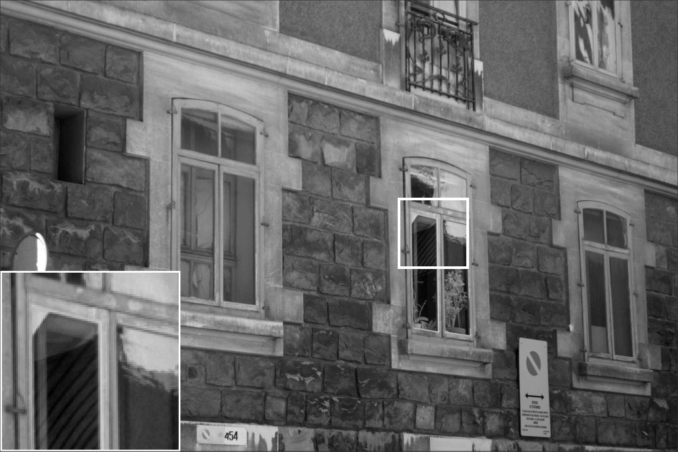}
	\end{minipage} 
	\begin{minipage}[b]{0.32\linewidth}
		\centering
		\includegraphics[width = 2.8cm, height= 2.1cm]{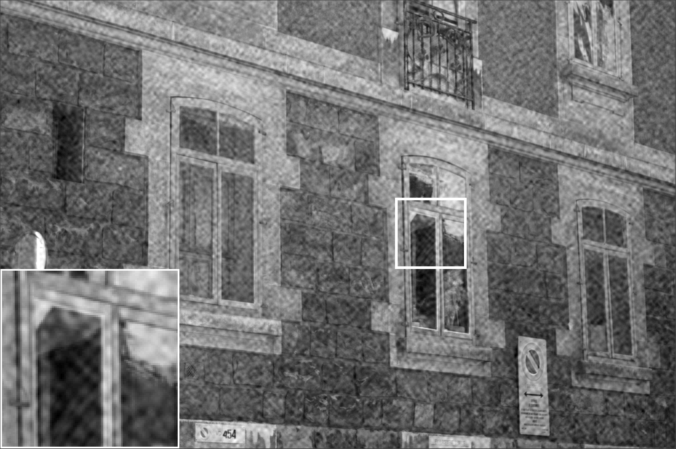}
	\end{minipage} 
	\begin{minipage}[b]{0.32\linewidth}
		\centering
		\includegraphics[width = 2.8cm, height= 2.1cm]{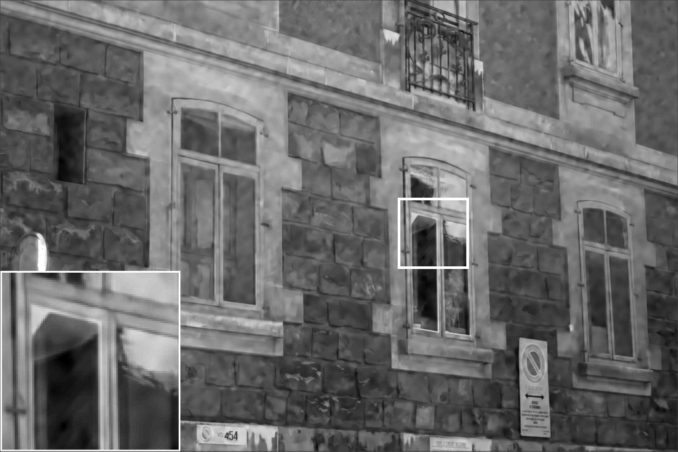}
	\end{minipage} 
	\\
	\begin{minipage}[b]{0.32\linewidth}
		\centering
		\includegraphics[width = 2.8cm, height= 2.1cm]{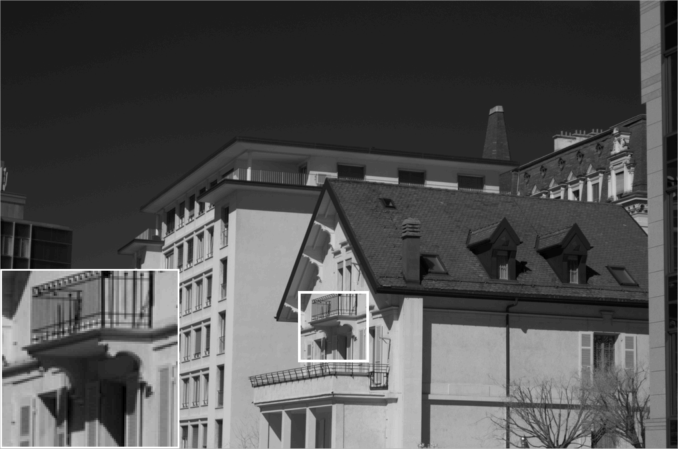}
	\end{minipage} 
	\begin{minipage}[b]{0.32\linewidth}
		\centering
		\includegraphics[width = 2.8cm, height= 2.1cm]{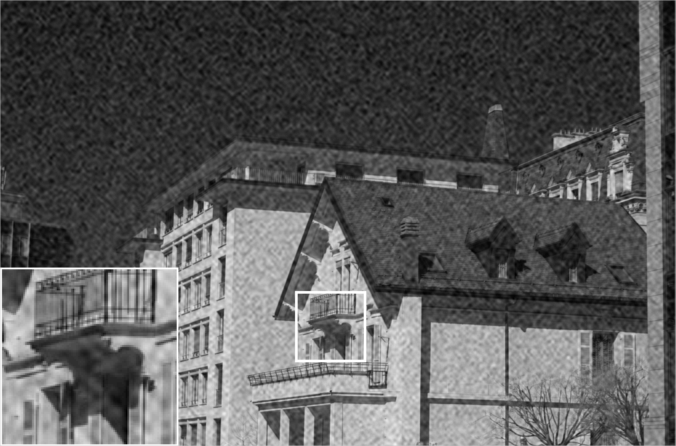}
	\end{minipage}  
	\begin{minipage}[b]{0.32\linewidth}
		\centering
		\includegraphics[width = 2.8cm, height= 2.1cm]{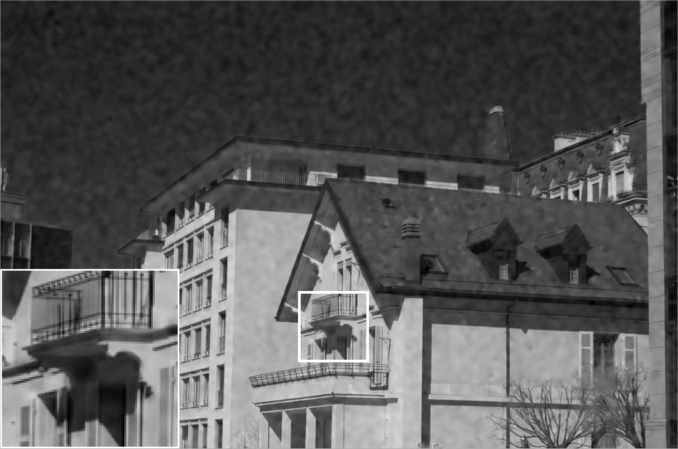}
	\end{minipage} 
	\\
	\begin{minipage}[b]{0.32\linewidth}
		\centering	{\footnotesize Truth} 
	\end{minipage} 
	\begin{minipage}[b]{0.32\linewidth}
		\centering	{\footnotesize DJF\cite{li2016deep}} 
	\end{minipage} 
	\begin{minipage}[b]{0.32\linewidth}
		\centering 	{\footnotesize Proposed}
	\end{minipage} 
	\caption{LR noisy testing images. The testing noise $\sigma_{test} = 12$ and the training noise $\sigma_{train} = 0$, for 4$\times$ upscaling of urban\_0006(up) and urban\_0020(bottom) using DJF~\cite{li2016deep} and proposed CDLSR.}
	\label{Fig:NIR4x_Joint_Noise}
\end{figure}

The first scenario assumes that the LR testing images are contaminated by zero-mean Gaussian noise with a certain standard deviation. Note that coupled dictionary learning is conducted on noiseless training images but coupled image super-resolution is conducted on the noisy LR images.
%
In Table~\ref{Tab:Noise} and Figure~\ref{Fig:NIR4x_Joint_Noise_Line}, the results corresponding to setting $\sigma_{test} = \sigma_{train} = 0$ show that deep-learning-based multimodal super-resolution method DJF\cite{li2016deep} usually outperforms our approach given a large number of training samples in the noise-free scenario. However, other results corresponding to setting $\sigma_{test} \neq 0$ show that our proposed algorithm demonstrates reasonable stability and robustness to noise, especially to strong noise. In contrast, DJF\cite{li2016deep} is susceptible to noise and its performance degrades faster than ours. In Figure~\ref{Fig:NIR4x_Joint_Noise}, it can be observed that the upscale results of DJF\cite{li2016deep} can not attenuate noise effectively, whereas our reconstruction is much cleaner. We believe that the good robustness and stability is due to sparsity priors exploited by our model.

\subsubsection{LR noisy both testing and training images}
\begin{figure}[t]
	\centering
	\begin{minipage}[b]{0.48\linewidth}
		\centering
		\includegraphics[width = 4.2cm]{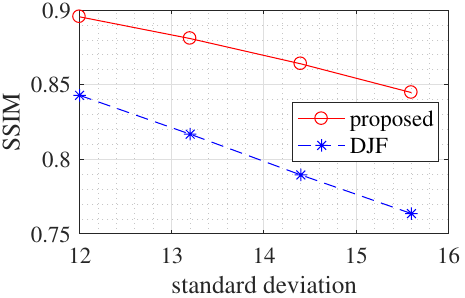} \\
		\scriptsize (a) SSIM w.r.t. Testing Noise
	\end{minipage} 
	\begin{minipage}[b]{0.48\linewidth}
		\centering
		\includegraphics[width = 4.2cm]{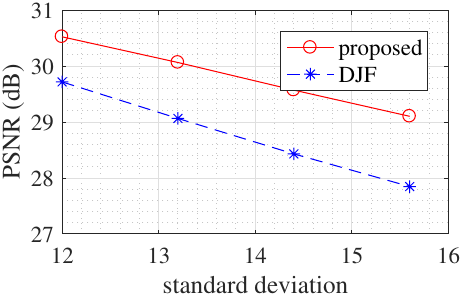} \\
		\scriptsize (b) PSNR w.r.t. Testing Noise
	\end{minipage}
	\caption{LR noisy both testing and training images. $\sigma_{train} = 12$ and $\sigma_{test}$ ranges from 12 to 15.6. 4$\times$ upscaling of near-infrared house images using DJF~\cite{li2016deep} and proposed CDLSR.}
	\label{Fig:NIR4x_Joint_NoiseMismatch_Line}
\end{figure}

The second scenario assumes that both the testing and the training images are contaminated by zero-mean Gaussian noise with a certain standard deviation. Coupled dictionary learning is performed on the noisy training images and coupled image super-resolution is done on the noisy testing image.
In addition, we consider possible \textit{mismatch} of noise in the LR testing and training images as well. Specifically, given a certain standard deviation $\sigma_{train}$ for the training noise and mismatch $\delta$, the standard deviation of the corresponding testing noise is set as $\sigma_{test} = \sigma_{train} (1+\delta).$ We add noise with standard deviation $\sigma_{train}$ in the LR training images and noise with standard deviation $\sigma_{test}$ in the LR testing images for various values of $\delta$. For example, given a typical noise level $\sigma_{train} = 12$ and mismatch $ \delta = [0, 10\%, 20\%, 30\%]$, it leads to corresponding $\sigma_{test} = [12, 13.2, 14.4, 15.6]$. Then we repeat the previous training and testing for 4$\times$ upscaling of near-infrared house images using both CDLSR and DJF\cite{li2016deep}.
As shown in Table~\ref{Tab:NIR4x_Joint_NoiseMismatch} and Figure~\ref{Fig:NIR4x_Joint_NoiseMismatch_Line}, the performance of both the proposed CDLSR approach and DJF\cite{li2016deep} degrades as the mismatch increases. However, the proposed algorithm not only has a slower degradation in performance than DJF\cite{li2016deep}, but also yields higher SSIM and PSNR values. This illustrates that our method is more robust to mismatched noise.
%

\begin{table*}[t]
	\centering
	\scriptsize
	\caption{LR noisy testing images. The standard deviation of the testing noise is $\sigma_{test} = [0, 4, 8, 12]$ while the standard deviation of the training noise is $\sigma_{train} = 0$, for 4$\times$ upscaling of near-infrared house images using DJF~\cite{li2016deep} and CDLSR.}
	\label{Tab:Noise}
	\begin{tabular}{C{1.2cm}|| L{0.55cm}L{0.55cm}| L{0.55cm}L{0.55cm}|| L{0.55cm}L{0.55cm}| L{0.55cm}L{0.55cm}|| L{0.55cm}L{0.55cm}| L{0.55cm}L{0.55cm}|| L{0.55cm}L{0.55cm}| L{0.55cm}L{0.55cm}}
		\hline \hline
		& \multicolumn{4}{c||}{$\sigma_{train} = 0$, $\sigma_{test}=0$} & \multicolumn{4}{c||}{$\sigma_{train} = 0$, $\sigma_{test}=4$} & \multicolumn{4}{c||}{$\sigma_{train} = 0$, $\sigma_{test}=8$} & \multicolumn{4}{c}{$\sigma_{train} = 0$, $\sigma_{test}=12$} \\
		\hline
		& \multicolumn{2}{c|}{Proposed} & \multicolumn{2}{c||}{DJF\cite{li2016deep}} & \multicolumn{2}{c|}{Proposed} & \multicolumn{2}{c||}{DJF\cite{li2016deep}} & \multicolumn{2}{c|}{Proposed} & \multicolumn{2}{c||}{DJF\cite{li2016deep}} & \multicolumn{2}{c|}{Proposed} & \multicolumn{2}{c}{DJF\cite{li2016deep}} \\
		& SSIM & PSNR & SSIM & PSNR & SSIM & PSNR & SSIM & PSNR & SSIM & PSNR & SSIM & PSNR & SSIM & PSNR & SSIM & PSNR \\
		\hline
		urban\_0004 & 0.9811 & 34.14 & \textbf{0.9895} & \textbf{35.85} & \textbf{0.9715} & 32.96 & 0.9386 & \textbf{33.62} & \textbf{0.9477} & \textbf{31.97} & 0.8313 & 30.21 & \textbf{0.9150} & \textbf{30.71} & 0.7181 & 27.42 \\
		urban\_0006 & 0.9868 & 36.79 & \textbf{0.9917} & \textbf{37.92} & \textbf{0.9752} & \textbf{35.77} & 0.9524 & 34.54 & \textbf{0.9483} & \textbf{33.71} & 0.8574 & 30.33 & \textbf{0.9132} & \textbf{31.87} & 0.7493 & 27.23 \\
		urban\_0017 & 0.9777 & 35.27 & \textbf{0.9861} & \textbf{37.00} & \textbf{0.9592} & \textbf{34.62} & 0.9233 & 34.19 & \textbf{0.9243} & \textbf{33.08} & 0.7896 & 30.30 & \textbf{0.8777} & \textbf{31.47} & 0.6571 & 27.33 \\
		urban\_0018 & 0.9874 & 33.01 & \textbf{0.9933} & \textbf{34.48} & \textbf{0.9801} & 32.58 & 0.9537 & \textbf{32.75} & \textbf{0.9594} & \textbf{31.88} & 0.8650 & 29.75 & \textbf{0.9302} & \textbf{30.72} & 0.7766 & 27.13 \\
		urban\_0020 & 0.9893 & 36.66 & \textbf{0.9953} & \textbf{38.27} & \textbf{0.9784} & \textbf{36.11} & 0.9334 & 34.94 & \textbf{0.9479} & \textbf{34.51} & 0.8070 & 30.74 & \textbf{0.9063} & \textbf{32.64} & 0.6807 & 27.63 \\
		urban\_0026 & 0.9482 & 30.35 & \textbf{0.9635} & \textbf{31.53} & \textbf{0.9339} & 30.00 & 0.9105 & \textbf{30.58} & \textbf{0.9104} & \textbf{29.46} & 0.7968 & 28.54 & \textbf{0.8788} & \textbf{28.69} & 0.6808 & 26.44 \\
		urban\_0030 & 0.9443 & 32.71 & \textbf{0.9604} & \textbf{35.10} & \textbf{0.9278} & 32.25 & 0.9101 & \textbf{33.09} & \textbf{0.9043} & \textbf{31.37} & 0.8003 & 29.81 & \textbf{0.8754} & \textbf{30.29} & 0.6907 & 27.06 \\
		urban\_0050 & \textbf{0.9663} & \textbf{29.37} & 0.9586 & 28.31 & \textbf{0.9465} & \textbf{28.92} & 0.8935 & 27.84 & \textbf{0.9138} & \textbf{28.54} & 0.7559 & 26.66 & \textbf{0.8687} & \textbf{28.01} & 0.6270 & 25.23 \\
		\hline
		average & 0.9726 & 33.54 & \textbf{0.9798} & \textbf{34.81} & \textbf{0.9591} & \textbf{32.90} & 0.9269 & 32.69 & \textbf{0.9320} & \textbf{31.82} & 0.8129 & 29.54 & \textbf{0.8956} & \textbf{30.55} & 0.6975 & 26.93 \\
		\hline \hline
	\end{tabular}
\end{table*}

\begin{table*}[t]
	\centering
	\scriptsize
	\caption{LR noisy both testing and training images. The standard deviation of the training noise is $\sigma_{train} = 12$ and the standard deviation of the testing noise $\sigma_{test}$ ranges from 12 to 15.6, corresponding to mismatch $\delta$ ranging from 0 to 30\%. 4$\times$ upscaling of near-infrared house images using DJF~\cite{li2016deep} and proposed CDLSR.}
	\label{Tab:NIR4x_Joint_NoiseMismatch}
	\begin{tabular}{C{1.2cm}|| L{0.55cm}L{0.55cm}| L{0.55cm}L{0.55cm}|| L{0.55cm}L{0.55cm}| L{0.55cm}L{0.55cm}|| L{0.55cm}L{0.55cm}| L{0.55cm}L{0.55cm}|| L{0.55cm}L{0.55cm}| L{0.55cm}L{0.55cm}}
		\hline \hline
		& \multicolumn{4}{c||}{$\sigma_{train} = 12$, $\sigma_{test} = 12$} & \multicolumn{4}{c||}{$\sigma_{train} = 12$, $\sigma_{test} = 13.2$} & \multicolumn{4}{c||}{$\sigma_{train} = 12$, $\sigma_{test} = 14.4$} & \multicolumn{4}{c}{$\sigma_{train} = 12$, $\sigma_{test} = 15.6$} \\
		\hline
		& \multicolumn{2}{c|}{Proposed} & \multicolumn{2}{c||}{DJF\cite{li2016deep}} & \multicolumn{2}{c|}{Proposed} & \multicolumn{2}{c||}{DJF\cite{li2016deep}} & \multicolumn{2}{c|}{Proposed} & \multicolumn{2}{c||}{DJF\cite{li2016deep}} & \multicolumn{2}{c|}{Proposed} & \multicolumn{2}{c}{DJF\cite{li2016deep}} \\
		& SSIM & PSNR & SSIM & PSNR & SSIM & PSNR & SSIM & PSNR & SSIM & PSNR & SSIM & PSNR & SSIM & PSNR & SSIM & PSNR \\
		\hline
		urban\_0004 & \textbf{0.9148} & \textbf{30.67} & 0.8724 & 30.38 & \textbf{0.9029} & \textbf{30.21} & 0.8545 & 29.84 & \textbf{0.8884} & \textbf{29.73} & 0.8233 & 29.03 & \textbf{0.8716} & \textbf{29.28} & 0.7981 & 28.38 \\
		urban\_0006 & \textbf{0.9131} & \textbf{31.82} & 0.8746 & 30.57 & \textbf{0.8973} & \textbf{31.17} & 0.8497 & 29.77 & \textbf{0.8789} & \textbf{30.48} & 0.8264 & 29.05 & \textbf{0.8585} & \textbf{29.84} & 0.8022 & 28.33 \\
		urban\_0017 & \textbf{0.8777} & \textbf{31.46} & 0.8141 & 30.63 & \textbf{0.8606} & \textbf{30.98} & 0.7792 & 29.71 & \textbf{0.8412} & \textbf{30.46} & 0.7474 & 29.07 & \textbf{0.8188} & \textbf{29.93} & 0.7156 & 28.40 \\
		urban\_0018 & \textbf{0.9301} & \textbf{30.70} & 0.8903 & 29.97 & \textbf{0.9174} & \textbf{30.20} & 0.8713 & 29.39 & \textbf{0.9029} & \textbf{29.67} & 0.8479 & 28.70 & \textbf{0.8870} & \textbf{29.19} & 0.8306 & 28.16 \\
		urban\_0020 & \textbf{0.9062} & \textbf{32.58} & 0.8442 & 31.30 & \textbf{0.8895} & \textbf{31.94} & 0.8161 & 30.50 & \textbf{0.8704} & \textbf{31.27} & 0.7824 & 29.66 & \textbf{0.8492} & \textbf{30.68} & 0.7608 & 29.04 \\
		urban\_0026 & \textbf{0.8786} & \textbf{28.66} & 0.8344 & 28.60 & \textbf{0.8661} & \textbf{28.36} & 0.8097 & 28.08 & \textbf{0.8511} & \textbf{28.02} & 0.7877 & 27.67 & \textbf{0.8341} & \textbf{27.73} & 0.7606 & 27.19 \\
		urban\_0030 & \textbf{0.8754} & \textbf{30.29} & 0.8275 & 29.83 & \textbf{0.8629} & \textbf{29.85} & 0.8007 & 29.09 & \textbf{0.8483} & \textbf{29.39} & 0.7785 & 28.44 & \textbf{0.8313} & \textbf{28.90} & 0.7498 & 27.76 \\
		urban\_0050 & \textbf{0.8687} & \textbf{28.01} & 0.7860 & 26.47 & \textbf{0.8514} & \textbf{27.79} & 0.7539 & 26.14 & \textbf{0.8311} & \textbf{27.55} & 0.7229 & 25.83 & \textbf{0.8076} & \textbf{27.27} & 0.6915 & 25.53 \\
		\hline
		average & \textbf{0.8956} & \textbf{30.52} & 0.8429 & 29.72 & \textbf{0.8810} & \textbf{30.07} & 0.8169 & 29.06 & \textbf{0.8640} & \textbf{29.57} & 0.7896 & 28.43 & \textbf{0.8448} & \textbf{29.10} & 0.7636 & 27.85 \\
		\hline \hline
	\end{tabular}
\end{table*}

\vspace{-0.3cm}

\subsection{Impact of parameters}
In this section, we illuminate further the performance of the CDLSR algorithm by exploring the effect of key parameters and factors on the performance of the proposed algorithms.
The following experiments are conducted using multi-spectral images of wavelength 640 nm on which we perform $4 \times$ upscaling using a computer equipped with a quadro-core i7 CPU at 3.4GHz with 32GB of memory. Each evaluation metric value is averaged on all the testing images.

\mypar{Dictionary Size}
Intuitively, more atoms tend to capture more features. Thus, a larger dictionary may yield a more accurate sparse approximation to the signal of interest. On the other hand, a large dictionary size increases the complexity of the non-convex problem, thus requiring more computation. Under the multi-spectral image SR experimental setting, we evaluate the performance of the proposed approach for various dictionary sizes, including 64, to 128, 256, 512, and 1024 atoms.
Table \ref{Tab:EffectDsize} shows that the PSNR increases gradually with the increase of the dictionary size. On the other hand, the computation cost, represented by the training time and testing time, approximately increases linearly with the dictionary size. The results imply that the choice of the dictionary size depends on the balance between approximation accuracy and computational expense. 



\begin{table}[t]
	\centering
	\caption{Effect of dictionary size}
	\begin{tabular}{l | llllll}
		\hline \hline
		\# of atoms & 64 & 128 & 256 & 512 & 1024 & 2048 \\
		\hline
		Train Time & \phantom{0}1.8 & \phantom{0}5.5 & 10.0 &	21.9 &	62.7 &	253.8 \\
		Test Time & 92.5 & 95.6 & 117.6 & 138.8 & 142.0 & 216.3 \\
		PSNR (dB) & 33.81 & 34.38 & 34.57 & 34.83 & 34.95 & 35.33 \\
		RMSE & 0.0204 & 0.0191 & 0.0187 & 0.0181 & 0.0179 & 0.0171 \\
		\hline \hline
	\end{tabular}
	\label{Tab:EffectDsize}
\end{table}

\mypar{Sparsity Constraints}
%
A larger sparsity constraint, i.e., a larger $s$, can lead to a better approximation of the data. On the other hand, larger sparsity constraints also require more iterations to find these non-zeros via OMP. As shown in Table~\ref{Tab:EffectSparsity}, the average PSNR of the reconstruction, as well as the computational time, increase along with the total sparsity constraint. When the total sparsity constraint goes beyond a certain level, e.g., 16, the retrieved extra non-zeros coefficients are trivial and contribute very little to the PSNR.

These considerations suggest that a dictionary size around 1024 atoms with sparsity constraint around 20 for $8 \times 8$ image patch size can yield decent performance while allowing affordable computational complexity.

\begin{table}[t]
	\centering
	\caption{Effect of sparsity constraints}
	\begin{tabular}{l | llllll ll}
		\hline \hline
		total sparsity & $8$ & $12$ & $16$ & $20$ & $24$ & $28$ \\
		\hline
		Test Time (s) &	\phantom{0}5.7 &	\phantom{0}8.8 &	12.6 &	17.9 & 23.2 & 31.9 \\
		PSNR (dB) &	34.97 &	35.54 &	35.69 &	35.73 & 35.73 & 35.69 \\
		\hline \hline
	\end{tabular}
	\label{Tab:EffectSparsity}
\end{table}

\vspace{-0.3cm}

\section{Conclusion}

\vspace{-0.2cm}

This paper proposed a new multimodal image SR approach based on joint sparse representations and coupled dictionary learning. In particular, our CDLSR approach explicitly captures the similarities and disparities between different image modalities in the sparse feature domain in \emph{lieu} of the image domain.
The proposed CDLSR approach consists of a training phase and a testing phase. The training phase seeks to learn a number of coupled dictionaries from training data and the testing phase leverages the learned dictionaries to reconstruct a HR version of a LR image with the aid of the guidance image. Our design automatically transfers appropriate structure information to the estimated HR version. Multispectral/RGB and NIR/RGB multimodal image SR experiments demonstrate that our design brings notable benefits over state-of-the-art image SR approaches. 
Our approach also outperforms deep-learning-based methods especially when the data is contaminated by noise, demonstrating better robustness, but consuming much less computing resource and training time.

\pagebreak


%


\clearpage

\end{document}